\documentclass[runningheads]{llncs}

\usepackage{amsmath}
\usepackage{amssymb}
\usepackage[accsupp]{axessibility}
\usepackage{bm}
\usepackage{booktabs}
\usepackage{enumitem}
\usepackage{graphicx}
\usepackage{hyperref}
\usepackage{siunitx}
\usepackage{wrapfig}

\newcommand{\paragraphskip}{\smallskip}
\newcommand{\paragraphtitle}[1]{\noindent \textbf{#1}}

\setlist{nolistsep}

\def\ECCVSubNumber{3751}

\titlerunning{Event Neural Networks}
\authorrunning{M. Dutson et al.}  
\author{Matthew~Dutson\orcidID{0000-0003-3133-2115} \and Yin~Li\orcidID{0000-0003-4173-9453} \and Mohit~Gupta\orcidID{0000-0002-2323-7700}}
\institute{
    University of Wisconsin--Madison, Madison WI 53715, USA \\
    \email{\{dutson,yin.li,mgupta37\}@wisc.edu}
}

\title{Event Neural Networks}

\begin{document}

\maketitle

\begin{abstract}
Video data is often repetitive; for example, the contents of adjacent frames are usually strongly correlated. Such redundancy occurs at multiple levels of complexity, from low-level pixel values to textures and high-level semantics. We propose Event Neural Networks (EvNets), which leverage this redundancy to achieve considerable computation savings during video inference. A defining characteristic of EvNets is that each neuron has state variables that provide it with long-term memory, which allows low-cost, high-accuracy inference even in the presence of significant camera motion. We show that it is possible to transform a wide range of neural networks into EvNets without re-training. We demonstrate our method on state-of-the-art architectures for both high- and low-level visual processing, including pose recognition, object detection, optical flow, and image enhancement. We observe roughly an order-of-magnitude reduction in computational costs compared to conventional networks, with minimal reductions in model accuracy.
\end{abstract}

\section{Introduction}
\label{sec:introduction}

\begin{figure}[t]
    \centering
    \includegraphics{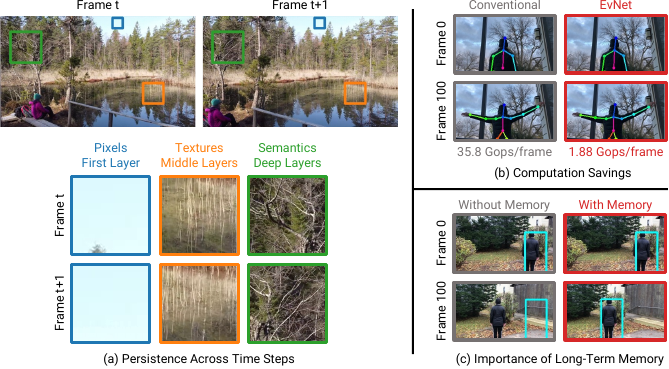}
    \caption{\textbf{Event Neural Networks.} \textbf{(a)} Two frames from a video sequence, separated by 1 second. Video source: \cite{sillerkiil2021Eesti}. Over this time, some areas of the image maintain consistent pixel values (sky region). However, these areas only represent a small fraction of the frame. In other regions, the pixel values change but the textures (vertical lines) or semantics (tree branches) remain the same. Each type of persistence corresponds to a different depth in the neural hierarchy. EvNets leverage temporal persistence in video streams across multiple levels of complexity. \textbf{(b)} EvNets yield significant computation savings while maintaining high accuracy. \textbf{(c)} Event neurons have state variables that encode long-term memory, allowing EvNets to perform robust inference even over long video sequences with significant camera motion. A network without long-term memory (left) fails to correctly track the object due to gradual error accumulation.}
    \label{fig:teaser}
\end{figure}

Real-world visual data is repetitive; that is, it has the property of \emph{persistence}. For example, observe the two frames in \autoref{fig:teaser}~(a). Despite being separated by one second, they appear quite similar. Human vision relies on the persistent nature of visual data to allocate limited perceptual resources. Instead of ingesting the entire scene at high resolution, the human eye points the fovea (a small region of dense receptor cells) at areas containing motion or detail~\cite{ware2008Visual}. This allocation of attention reduces visual processing and eye-to-brain communication.

Processing individual frames using artificial neural networks has proven to be a competitive solution for video inference~\cite{xiao2018simple,zhu2018High}. This paradigm leverages advances in \emph{image} recognition (e.g., pose estimation or object detection) and processes each frame independently without considering temporal continuity, implicitly assuming that adjacent frames are statistically independent. This assumption leads to inefficient use of resources due to the repeated processing of image regions containing little or no new information.

There has been recent interest in leveraging temporal redundancy for efficient video inference. One simple solution is to skip processing image regions containing few changes in pixel values. However, such methods cannot recognize persistence in textures, patterns, or high-level semantics when it does not coincide with persistent pixel values. See \autoref{fig:teaser}~(a).

Because neural networks extract a hierarchy of features from their inputs, they contain a built-in lens for detecting repetition across many levels of visual complexity. Shallow layers detect low-level patterns, and deep layers detect high-level semantics. Temporal repetition at a given level of complexity translates to persistent values at the corresponding depth in the neural hierarchy~\cite{habibian2021Skipconvolutions}. Based on this observation, we propose Event Neural Networks (EvNets), a family of neural networks in which neurons transmit (thereby triggering downstream computation) only when there is a significant change in their activation. By applying this strategy over all neurons and layers, we detect and exploit temporal persistence across many levels of complexity.

One of the defining features of EvNets is that each neuron has state variables that provide it with \emph{long-term memory}. Instead of re-computing from scratch for every new input, an EvNet neuron accumulates information over time. Long-term memory allows EvNets to perform robust inference over long video sequences containing significant camera motion. See \autoref{fig:teaser}~(c).

We design various structural components for EvNets -- both at the individual neuron level (memory state variables) and the network level (layers and transmission policies). We recognize that transmission policies, in particular, are critical for achieving a good accuracy/computation tradeoff, and we describe the policy design space in detail. We show that, with these components, it is possible to transform a broad class of conventional networks into EvNets \emph{without re-training}. We demonstrate our methods on state-of-the-art models for several high- and low-level tasks: pose recognition, object detection, optical flow, and image enhancement. We observe approximately an order-of-magnitude reduction in arithmetic operations with minimal effects on model accuracy.

\paragraphskip
\paragraphtitle{Scope and Limitations.} In this paper, we focus on the theoretical and conceptual properties of EvNets. Although we show results on several video inference tasks, our goal is not to compete with the latest methods for these tasks in terms of accuracy. Instead, we show that, across a range of models and tasks, EvNets can significantly reduce computational costs without decreasing accuracy.

In most of our analyses we do not assume a specific hardware platform or computation model. We mainly report arithmetic and memory operations (a platform-invariant measure of computational cost) instead of wall-clock time (which depends on many situational variables). An important next step is to consider questions relating to the design of hardware-software stacks for EvNets, in order to minimize latency and power consumption.

\section{Related Work}
\label{sec:related_work}

\paragraphtitle{Efficient Neural Networks.} There are numerous methods for reducing the computational cost of neural networks. Many architectures have been designed to require fewer parameters and arithmetic operations~\cite{howard2017MobileNets,iandola2016SqueezeNet,lin2017Focal,liu2016SSD,redmon2016You,zhang2018ShuffleNet}. Another line of work uses low-precision arithmetic to achieve computation savings~\cite{courbariaux2015BinaryConnect,hwang2014Fixedpoint,rastegari2016XNORNet,vanhoucke2011Improving}. Our approach is complementary to both architecture- and precision-based efficiency methods. These methods reduce the cost of inference on a single time step, whereas EvNets eliminate repetitive computation between multiple time steps. Pruning algorithms~\cite{han2015Learning,hassibi1992Second,lecun1989Optimal,li2017Pruning} remove redundant neurons or synapses during training to improve efficiency. Instead of pruning universally redundant neurons, an EvNet adaptively ignores temporally redundant neurons.

\paragraphskip
\paragraphtitle{Adaptive Networks.} Adaptive models modify their computation based on the input to suit the difficulty of each inference. Prior approaches consider an ensemble of sub-networks~\cite{huang2016deep,veit2018Convolutional}, equip a network with multiple exits~\cite{huang2017multi,teerapittayanon2016BranchyNet}, select the input resolution at inference time~\cite{chin2019AdaScale,meng2020ar,yang2020Resolution}, or dynamically choose the feature resolution~\cite{wu2019liteeval}. These methods are designed for image recognition tasks and do not explore temporal redundancy. Further, many require custom tailoring or re-training for each task and architecture. In contrast, EvNets can be readily integrated into many existing architectures and do not require re-training.

\paragraphskip
\paragraphtitle{Temporal Redundancy.} Several recent approaches consider temporal redundancy for efficient video inference. Many take a keyframe-oriented approach, computing expensive features on keyframes, then transforming those features for the other frames~\cite{chen2020memory,jain2019Accel,li2018Lowlatency,shelhamer2016Clockwork,zhu2018High,zhu2017Deep}. Other methods include using visual trackers~\cite{xu2020ApproxDet}, skipping redundant frames~\cite{ghodrati2021frameexit,wu2019adaframe}, reusing previous frame features~\cite{meng2021adafuse}, distilling results from previous time steps~\cite{nie2019Dynamic}, two-stream computation~\cite{feichtenhofer2019SlowFast}, and leveraging video compression~\cite{wu2018Compressed}. In general, these methods require extensive modifications to the network architecture.

Skip-convolution networks (Skip-Conv)~\cite{habibian2021Skipconvolutions} are closely related to EvNets. Skip-Conv reuses activation values that have not changed significantly between frames. However, the algorithm only tracks changes between consecutive frames and thus requires frequent re-initialization to maintain accuracy. Re-initialization leads to reduced efficiency, especially in the presence of camera motion. In contrast, the long-term memory in an EvNet maintains accuracy and efficiency over hundreds of frames, even when there is strong camera motion. See \autoref{fig:teaser}~(c).

Sigma-Delta networks~\cite{oconnor2016Sigma} exploit temporal redundancy by quantizing the changes in neuron activations. Sigma-Delta networks have been limited so far to simple tasks like digit classification. Unlike Sigma-Delta networks, EvNets do not require quantization (although they do allow it). Compared to Sigma-Delta networks, EvNets achieve superior accuracy/computation tradeoffs (\autoref{fig:policy_design}) and generalize better to challenging, real-world tasks (\autoref{fig:sample_outputs}).

DeltaCNN~\cite{parger2022DeltaCNN} is concurrent work with similar goals to this paper. Like EvNets, DeltaCNN models have mechanisms for integrating long-term changes. They focus on translating theoretical speedups into GPU wall-time savings by enforcing structured sparsity (all channels at a given location transmit together). Despite its practical benefits, this design is inefficient when there is camera motion. In contrast, we emphasize broad conceptual frameworks (e.g., arbitrary sparsity structure) with an eye toward future hardware architectures (\autoref{sec:discussion}).

\paragraphskip
\paragraphtitle{Event Sensor Inference.} Event sensors~\cite{lichtsteiner2008128} generate sparse frames by computing a quantized temporal gradient at each pixel. Many networks designed for inference on event-sensor data have efficient, sparse dynamics \cite{cannici2019Asynchronous,messikommer2020Eventbased}. However, they make strong assumptions about the mathematical properties of the network (e.g., that it is piecewise linear~\cite{messikommer2020Eventbased}). EvNets place far fewer constraints on the model and are compatible with a broad range of existing architectures.

\paragraphskip
\paragraphtitle{Recurrent Neural Networks (RNNs).} EvNets use long-term memory to track changes, and are thus loosely connected to RNNs. Long-term memory has been widely adopted in RNNs~\cite{hochreiter1997long}. Several recent works also propose adaptive inference for RNNs by learning to skip state updates~\cite{campos2017skip} or updating state variables only when a significant change occurs~\cite{neil2017Delta,pan2018Recurrent}. Unlike EvNets, these approaches are tailored for RNNs and generally require re-training.

\section{Event Neurons}
\label{sec:event_neurons}

\begin{figure}[t]
    \centering
    \includegraphics{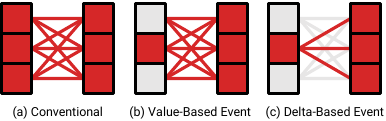}
    \caption{\textbf{Sparse, Delta-Based Transmission.} \textbf{(a)} Conventional neurons completely recompute their activations on each time step. \textbf{(b)} Value-based event neurons only transmit activations that have changed significantly. However, a value-based transmission can still trigger many computations. \textbf{(c)} Delta-based event neurons only transmit differential updates to their activations.}
    \label{fig:neuron_transmission}
\end{figure}

Consider a neuron in a conventional neural network. Let $\bm{x} = [x_1, x_2, \ldots, x_n]$ be the vector of input values and $y$ be the output. Suppose the neuron composes a linear function $g$ (e.g., a convolution) with a nonlinear activation $f$. That is,
\begin{equation}
\label{eq:linear_function}
	\textstyle g(\bm{x}) = \sum_{i=1}^n w_i x_i; \quad y = f(g(\bm{x})),
\end{equation}
where $\bm{w} = [w_1, w_2, \ldots, w_n]$ contains the weights of the function $g$. In a conventional network, every neuron recomputes $f$ and $g$ for every input frame (\autoref{fig:neuron_transmission}~(a)), resulting in large computational costs over a video sequence.

Inspired by prior methods that exploit persistence in activations \cite{campos2017skip,habibian2021Skipconvolutions,oconnor2016Sigma}, we describe a class of \emph{event neurons} with \emph{sparse}, \emph{delta-based} transmission.

\paragraphskip
\paragraphtitle{Sparse, Delta-Based Transmission.} An event neuron transmits its output to subsequent layers only when there is a sufficient change between its current activation and the previous transmission. This gating behavior makes the layer output \emph{sparse}. However, a \emph{value} transmission may still trigger many downstream computations (neurons receiving updated input values must recompute their activations from scratch). See \autoref{fig:neuron_transmission}~(b). Therefore, instead of transmitting an activation \emph{value}, an event neuron transmits a \emph{delta} (differential).

Suppose a neuron receives a vector of incoming differentials $\bm{\Delta}_\text{in}$ (one element per incoming synapse). $\bm{\Delta}_\text{in}$ is sparse, i.e., it only contains nonzero values for upstream neurons that have transmitted. The updated $g$ is given by
\begin{equation}
    g(\bm{x} + \bm{\Delta}_\text{in}) = g(\bm{x}) + g(\bm{\Delta}_\text{in}).
\end{equation}
Instead of computing $g(\bm{x} + \bm{\Delta}_\text{in})$ from scratch, an event neuron stores the value of $g(\bm{x})$ in a state variable $a$. When it receives a new input $\bm{\Delta}_\text{in}$, the neuron retrieves the old value of $g(\bm{x})$ from $a$, computes $g(\bm{\Delta}_\text{in})$, and saves the value $g(\bm{x}) + g(\bm{\Delta}_\text{in})$ in $a$. This process only requires computing products $w_i x_i$ for nonzero elements of $\bm{\Delta}_\text{in}$, i.e., computation scales linearly with the number of transmissions.

The activation function $f$ is nonlinear, so we cannot update it incrementally like $g$. Whenever $a$ changes, we recompute $f(a)$, then store the updated value in another state variable. $f$ is usually a lightweight function (e.g., ReLU), so the cost of recomputing $f$ is far smaller than computing the products $w_i x_i$.

\subsection{Building Event Neurons}
\label{sec:building_event_neurons}

\begin{figure}[t]
    \centering
    \includegraphics{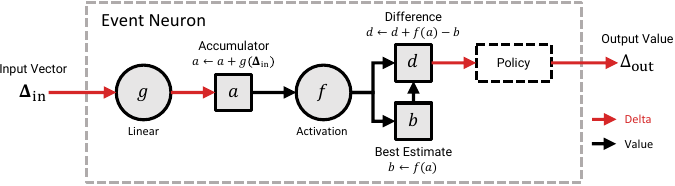}
    \caption{\textbf{Building Event Neurons.} The state variables and update rules in an event neuron. The incremental updates to $a$ convert from a delta-based representation to value-based. The subtraction $f(a) - b$ returns the output to delta-based.}
    \label{fig:building_event_neurons}
\end{figure}

An event neuron consists of three state variables, as shown in \autoref{fig:building_event_neurons}. The \emph{accumulator} ($a$) stores the current value of $g(\bm{x})$. The \emph{best estimate} ($b$) stores the current value of $f(a)$. The \emph{difference} ($d$) stores difference between $b$ and the most recent output. When a neuron receives a differential update $\bm{\Delta}_\text{in}^{(t)}$ at time $t$ from one or more of its inputs, it updates these state variables as follows:
\begin{equation}
\label{eq:neuron_update}
    a^{(t + 1)} = a^{(t)} + g(\bm{\Delta}_\text{in}^{(t)}); \quad d^{(t + 1)} = d^{(t)} + f(a^{(t + 1)}) - b^{(t)}; \quad b^{(t + 1)} = f(a^{(t + 1)}).
\end{equation}

A neuron transmits an output $\Delta_\text{out}$ when some condition on $d$ is satisfied. This condition is defined by the \emph{transmission policy}. The transmission policy also gives the relationship between $d$ and $\Delta_\text{out}$. The policies in this paper simply set $\Delta_\text{out} = d$. However, other relationships are possible, and the properties described in \autoref{sec:properties_of_event_neurons} hold for other relationships. After a neuron transmits, it sets $d$ to $d - \Delta_\text{out}$. See \autoref{sec:transmission_policies} for more details on transmission policies.

\subsection{Properties of Event Neurons} 
\label{sec:properties_of_event_neurons}

\paragraphtitle{Long- and Short-Term Memory.} The state variable $d$ accumulates all not-yet-transmitted corrections to the neuron output. It represents the neuron's \emph{long-term memory}, whereas $b$ represents its \emph{short-term memory}. Including a long-term memory keeps the neuron from discarding information when it does not transmit. This \emph{error-retention property} grants certain guarantees on the neuron's behavior, as we demonstrate next.

\paragraphskip
\paragraphtitle{Error Retention.} Consider an event neuron receiving a series of inputs over $T$ time steps, $\bm{\Delta}_\text{in}^{(1)}, \bm{\Delta}_\text{in}^{(2)}, \ldots, \bm{\Delta}_\text{in}^{(T)}$. Assume that the state variables $a$, $b$, and $d$ have initial values $a^{(0)}$, $f(a^{(0)})$, and zero, respectively. Let the transmitted output values at each time step be $\Delta_\text{out}^{(1)}, \Delta_\text{out}^{(2)}, \ldots, \Delta_\text{out}^{(T)}$ (some of these may be zero). By repeatedly applying the neuron update rules, we arrive at the state
\begin{equation}
\begin{gathered}
    \textstyle a^{(T)} = a^{(0)} + g \left( \sum_{t=1}^T \bm{\Delta}_\text{in}^{(t)} \right); \quad b^{(T)} = f(a^{(T)}); \\
    \textstyle d^{(T)} = b^{(T)} - b^{(0)} - \sum_{t=1}^T \Delta_\text{out}^{(t)}. 
\end{gathered}
\end{equation}
See the supplementary material for a detailed derivation. Observe that $d$ is equal to the difference between the actual and transmitted changes in the activation. This is true regardless of the order or temporal distribution of the $\bm{\Delta}_\text{in}$ and $\Delta_\text{out}$. Because the neuron stores $d$, it always has enough information to bring the transmitted activation into exact agreement with the true activation $b$. We can use this fact to bound the error within an EvNet. For example, we can constrain each neuron's error to the range $[-h, +h]$ by transmitting when $|d| > h$.

\paragraphskip
\paragraphtitle{The Importance of Long-Term Memory.} For comparison, consider a model in which neurons compute the difference between $b$ on adjacent time steps, then either transmit or discard this difference without storing the remainder. This is the model used in Skip-Conv~\cite{habibian2021Skipconvolutions}. Under this model, the final state of a neuron depends strongly on the order and temporal distribution of inputs.

For example, suppose a neuron transmits if the frame-to-frame difference exceeds a threshold $\delta$. Consider a scenario where the neuron's activation gradually increases from \num{0} to $2 \delta$ in steps $0.1 \delta, 0.2 \delta, \ldots, 2 \delta$. Gradual changes like this are common in practice (e.g., when panning over a surface with an intensity gradient). Because $0.1 \delta < \delta$, the neuron never transmits and ends in a state with error $-2 \delta$. The neuron carries this error into all of its future computations. Furthermore, because the neuron discards non-transmitted activations, it has no way to know that this $-2 \delta$ error exists.

\section{Event Networks}
\label{sec:event_networks}

\subsection{Building Event Networks}
\label{sec:building_event_networks}

So far, we have considered the design and characteristics of individual event neurons. In this section, we broaden our view and consider layers and networks. A ``layer'' is an atomic tensor operation (e.g., a convolution). By this definition, $g$ and $f$ as defined in \autoref{sec:building_event_neurons} correspond to two different layers.

We define three new layer types. An \emph{accumulator layer} consists of a state vector $\bm{a}$ that contains the $a$ variables for a collection of neurons. A \emph{gate layer} contains state vectors $\bm{b}$ and $\bm{d}$ and the transmission policy. A \emph{buffer layer} stores its inputs in a state vector $\bm{x}$ for future use by the next layer; this is required before non-pointwise, nonlinear layers like max pooling. The state vectors $\bm{a}$, $\bm{b}$ and $\bm{d}$ are updated using vectorized versions of the rules in \autoref{eq:neuron_update}. An accumulator layer converts its input from delta-based to value-based, whereas a gate converts from value-based to delta-based.

\begin{figure}[t]
    \centering
    \includegraphics{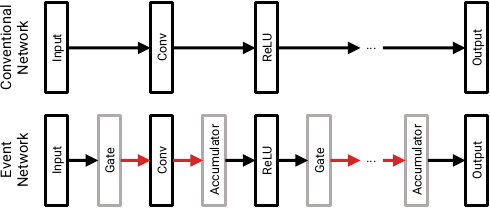}
    \caption{\textbf{Building Event Networks.} We insert accumulators and gates to make the input to linear layers (e.g., convolutions, fully-connected layers) delta-based and the input to nonlinear layers (e.g., ReLU activations) value-based.}
    \label{fig:building_event_networks}
\end{figure}

To create an EvNet, we insert gates and accumulators into a pretrained network such that linear layers receive delta inputs and nonlinear layers receive value inputs (\autoref{fig:building_event_networks}). Note that residual connections do not require any special treatment -- in an EvNet, residual connections simply carry deltas instead of values. These deltas are added or concatenated to downstream deltas when the residual branch re-joins the main branch (like in a conventional network).

We place a gate at the beginning of the network and an accumulator at the end. At the input gate, we use pixel values instead of $f(a)$ and update $b$ and $d$ at every timestep. At the output accumulator, we update $\bm{a}$ sparsely but read all its elements at every frame. Throughout the model, the functions computed by the preexisting layers (the $f$ and $g$) remain the same.

\subsection{Network Initialization}
\label{sec:network_initialization}

The equations in \autoref{sec:building_event_neurons} define how to update the neuron state variables, but they do not specify those variables’ initial values. Consider a simple initialization strategy where $a = 0$ and $d = 0$ for all neurons. Since the activation function $f$ is nonlinear, the value of the state variable $b = f(a)$ may be nonzero. This nonzero $b$ usually translates to a nonzero value of $a$ in the next layer. However, we initialized $a = 0$ for all neurons. We have an inconsistency.

To address this problem, we define the notion of \emph{internal consistency}. Consider a neuron with state variables $a$, $d$, and $b$. Let $\bm{b}_\text{in}$ and $\bm{d}_\text{in}$ be vectors containing the states of the neurons in the previous layer. We say that a network is in an internally consistent state if, for all neurons,
\begin{equation}
    a = g(\bm{b}_\text{in} - \bm{d}_\text{in}); \quad b = f(a).
\end{equation}
The simplest way to satisfy these criteria is to flush some canonical input through the network. Starting with neurons in the first layer and progressively moving through all subsequent layers, we set $a = g(\bm{b}_\text{in})$, $b = f(a)$, and $d = 0$. In our experiments, we use the first input frame as the canonical input.

\subsection{Transmission Policies}
\label{sec:transmission_policies}

A \emph{transmission policy} defines a pair of functions $M(\bm{d})$ and $P(\bm{d})$ for each layer. $M$ outputs a binary mask $\bm{m}$ indicating which neurons should transmit. $P$ outputs the values of $\bm{\Delta}_\text{out}$. In this subsection, we describe the transmission policy design space. The choice of transmission policy is a critical design consideration, strongly influencing the accuracy and efficiency of the final model.

\paragraphskip
\paragraphtitle{Locality and Granularity.} Policies may have different levels of \emph{locality}, defined as the number of elements from $\bm{d}$ required to compute each element of $\bm{m}$ and $\bm{\Delta}_\text{out}$. A \emph{global} policy considers all elements of $\bm{d}$ when computing each value $m_i$ and $\Delta_{\text{out},i}$. A \emph{local} policy considers some strict subset of $\bm{d}$, and an \emph{isolated} policy considers only the element $d_i$.

In addition to its locality, each policy has a \emph{granularity}. The granularity defines how $m$-values are shared between neurons. A \emph{chunked} policy ties neurons together into local groups, producing one value of $m$ for each group. Neurons in the same group fire in unison. This might be practically desirable for easy parallelization on the hardware. In contrast, a \emph{singular} policy assigns every neuron a separate value of $m$, so each neuron fires independently.

\paragraphskip
\paragraphtitle{A Linear-Cost Policy.} In this work, we use an isolated, singular policy based on a simple threshold. Specifically,
\begin{equation}
    m_i = H(|d_i| - h_i); \quad \Delta_{\text{out},i} = d_i,
\end{equation}
where $H$ is the Heaviside step function and $h_i$ is the threshold for neuron $i$. A key advantage of this policy is its low overhead. On receiving an incoming transmission, a neuron evaluates $|d| > h$ (one subtraction) in addition to the usual updates to $a$, $d$, and $b$. Neurons not receiving any updates (e.g., those in a static image region) do not incur any overhead for policy computations. In other words, the policy's cost is linear in the number of updated neurons. Combined with the linear cost of computing the neuron updates, this results in EvNets whose overall \emph{cost scales linearly with the amount of change in the input, not with the quantity of input data received}.

This linear cost has important implications for networks processing data from high-speed sensors (e.g., event~sensors~\cite{lichtsteiner2008128} or single-photon sensors~\cite{dutton2016SPADbased}). Here, the differences between adjacent inputs are often minuscule, and the cost of a policy with fixed per-frame overhead (e.g., a Gumbel~gate~\cite{habibian2021Skipconvolutions}) could come to dominate the runtime. EvNets with a linear-overhead policy are a natural solution for processing this type of high-speed data.

\begin{figure}[t]
    \centering
    \includegraphics{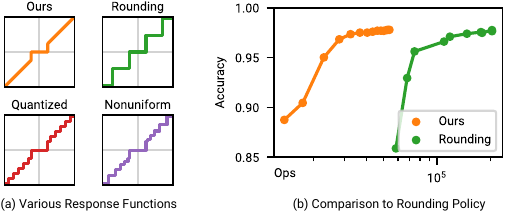}
    \caption{\textbf{Policy Design and Quantization.} \textbf{(a)} A few sample response functions (assuming an isolated, singular policy). \textbf{(b)} A comparison between our policy and a rounding policy (used in Sigma-Delta~networks~\cite{oconnor2016Sigma}). Results are for a 3-layer fully-connected network on the Temporal~MNIST~dataset~\cite{oconnor2016Sigma}.}
    \label{fig:policy_design}
\end{figure}

\paragraphskip
\paragraphtitle{Policy Design and Quantization.} When a policy is both isolated and singular, we can characterize the functions $M(\bm{d})$ and $P(\bm{d})$ by scalar functions $M(d_i)$ and $P(d_i)$. Taking the product $M(d_i) \cdot P(d_i)$ gives a \emph{response function} $R(d_i)$ that describes the overall behavior of the neuron. \autoref{fig:policy_design}~(a) illustrates several possible response functions.

Some response functions employ quantization to reduce the cost of computing dot product terms $w_i x_i$ (\autoref{eq:linear_function}). Sigma-Delta~networks~\cite{oconnor2016Sigma} use a rounding policy to quantize neuron outputs; a neuron transmits if this rounded value is nonzero. This rounding policy has significantly worse accuracy-computation tradeoffs (\autoref{fig:policy_design}~(b)) compared to our proposed policy. This might be caused by coupling the firing threshold with the quantization scale. To increase its output precision a Sigma-Delta network must reduce its firing threshold, possibly resulting in unnecessary transmissions.

\section{Experiments and Results}
\label{sec:experiments}

EvNets are widely applicable across architectures and video inference tasks. Any network satisfying a few basic requirements (i.e., frame-based and composing linear functions with nonlinear activations) can be converted to an EvNet \emph{without re-training}. To demonstrate this, we select widespread, representative models for our main experiments: YOLOv3~\cite{redmon2016You} for video object detection and OpenPose~\cite{cao2017Realtime} for video pose estimation. Additionally, we conduct ablation experiments and report results on low-level tasks (optical flow and image enhancement).

In the supplement, we include additional results on HRNet~\cite{sun2019deep} for pose estimation. We also include ablations for the effect of layer depth on computation cost, variations in computation over time, the effect of granularity on savings, improved temporal smoothness, and a comparison to simple interpolation. 

\subsection{Video Pose Estimation}
\label{sec:video_pose_estimation}

\paragraphtitle{Dataset and Experiment Setup.} We conduct experiments on the JHMDB dataset~\cite{jhuang2013Understanding} using the widely adopted OpenPose model~\cite{cao2017Realtime}. We use weights pre-trained on the MPII~dataset~\cite{andriluka20142D} from~\cite{cao2017Realtime} and evaluate the models on a subset of JHMDB with \num{319} videos and over \num{11}k frames, following~\cite{habibian2021Skipconvolutions}. We report results on the combination of the three JHMDB test splits. We use the PCK~metric~\cite{yang2013Articulated} with a detection threshold of $\alpha = 0.2$, consistent with prior works~\cite{habibian2021Skipconvolutions}.

\paragraphskip
\paragraphtitle{Implementation Details.} We resize all videos to $\num{320} \times \num{240}$, padding as needed to preserve the aspect ratio of the content. The joint definitions in MPII (the training dataset for OpenPose) differ slightly from those in JHMDB. During evaluation, we match the JHMDB ``neck,'' ``belly,'' and ``face'' joints to the MPII ``upper neck,'' ``pelvis,'' and ``head top'' joints, respectively.

\paragraphskip
\paragraphtitle{Baselines.} We consider the following baselines, all using the OpenPose model.
\begin{itemize}
    \item \textbf{Conventional}: This is the vanilla OpenPose model without modifications.
    \item \textbf{Skip-Conv}: This is a variant of the Skip-Conv method with norm gates and without periodic state resets.
    \item \textbf{Skip-Conv-8}: This adds state resets to Skip-Conv by re-flushing every \num{8} frames to reduce the effect of long-term activation drift.
\end{itemize}
We recognize that Skip-Conv networks can also incorporate a learnable gating function (the Gumbel gate) that uses information from a local window around each neuron. This can also be used for our EvNets (it is local and chunked rather than isolated and singular), but it requires re-training of the network and can incur a higher computational overhead. To keep the analysis fair, we only compare to the Skip-Conv norm gate.

\begin{figure}[t]
    \centering
    \includegraphics{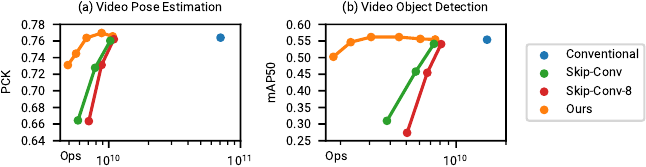}
    \caption{\textbf{Pareto Curves.} The performance of an EvNet over several different thresholds, with baselines for comparison. The ``Skip-Conv-8'' model re-flushes the network every \num{8} frames. EvNets give significant computation savings without sacrificing accuracy. See the supplementary material for a table with this data.}
    \label{fig:pareto_curves}
\end{figure}

\paragraphskip
\paragraphtitle{Results.} \autoref{fig:pareto_curves}~(a) presents our results. We vary the policy threshold $h$ to characterize the accuracy/computation Pareto frontier. For both Skip-Conv and EvNets, increasing the threshold reduces the computational cost but increases the error rate. EvNets consistently outperform their direct competitors (Skip-Conv and Skip-Conv reset) on the Pareto frontier, achieving significantly higher accuracy when using a similar amount of computation. Surprisingly, compared to the conventional OpenPose model, EvNets sometimes have slightly better accuracy, even with a large reduction in computation. We hypothesize that this is caused by a weak inter-frame ensembling effect.

\autoref{tab:accuracy_and_computation} summarizes the accuracy and computation at the best operating point on the Pareto curve. For each model, we choose the highest threshold that reduces PCK by less than \SI{0.5}{\percent}. To better understand the accuracy-computation tradeoff, we further report the compute and memory overhead of our EvNets (at the best operating point) in \autoref{tab:overhead}. We report overhead operations both as a number of operations and as a percentage. This percentage gives the ratio between ``extra operations expended'' and ``number of arithmetic operations saved.'' For example, an arithmetic overhead of \SI{0.12}{\percent} indicates that the neuron updates and transmission policy require \num{0.12} extra arithmetic operations for every \num{100} operations saved. Overall, EvNets add minimal operation overhead and manageable additional memory.

\begin{table}[t]
    \centering
    \caption{\textbf{Accuracy and Computation.} Results on the best threshold for each model. We choose the highest threshold that reduces PCK or mAP by less than \SI{0.5}{\percent}. See the supplement for a complete list of the points shown in \autoref{fig:pareto_curves}.}
    \label{tab:accuracy_and_computation}
    \begin{tabular}{lcccccc}
        \toprule
                     & \multicolumn{3}{c}{Pose Estimation}                  & \multicolumn{3}{c}{Object Detection}                 \\
                       \cmidrule(lr){2-4}                                     \cmidrule(lr){5-7}
        Model        & Thresh.    & PCK (\%)     & Operations               & Thresh.    & mAP (\%)     & Operations               \\ \midrule
        Conventional & --         & $\bm{76.40}$ & \num{7.055e10}           & --         & \num{55.38}  & \num{1.537e10}           \\
        Skip-Conv    & \num{0.01} & \num{76.03}  & \num{1.027e10}           & \num{0.01} & \num{54.13}  & \num{7.340e9}            \\
        Skip-Conv-8  & \num{0.01} & \num{76.21}  & \num{1.092e10}           & \num{0.01} & \num{54.06}  & \num{8.111e9}            \\
        EvNet        & \num{0.04} & \num{76.37}  & $\bm{6.780 \times 10^9}$ & \num{0.08} & $\bm{56.19}$ & $\bm{3.061 \times 10^9}$ \\
        \bottomrule
    \end{tabular}
\end{table}

\begin{table}[t]
    \centering
    \caption{\textbf{Overhead.} ``Weights'' gives the amount of memory required for model weights. ``Variables'' gives the amount of memory required for the state variables $a$, $b$, and $d$. ``Arithmetic'' indicates the number of extra arithmetic operations expended for neuron updates (\autoref{eq:neuron_update}) and policy-related computations. ``Load and Store'' indicates the number of extra memory access operations. See the text for an explanation of the percentage notation.}
    \label{tab:overhead}
    \begin{tabular}{lccccc}
        \toprule
                             &                  & \multicolumn{2}{c}{Memory Costs} & \multicolumn{2}{c}{Operation Overhead}    \\
                                                   \cmidrule(lr){3-4}                 \cmidrule(lr){5-6}
        Model    & Thresh.    & Weights        & Variables       & Arithmetic          & Load and Store          \\ \midrule
        OpenPose & \num{0.04} & \num{206.8} MB & \num{346.2} MB  & \num{7.570e7} (\SI{0.12}{\percent}) & \num{1.342e8} (\SI{0.21}{\percent}) \\
        YOLO     & \num{0.08} & \num{248.8} MB & \num{232.2} MB  & \num{6.417e7} (\SI{0.52}{\percent}) & \num{1.040e8} (\SI{0.85}{\percent}) \\
        \bottomrule
    \end{tabular}
\end{table}

\subsection{Video Object Detection}
\label{sec:video_object_detection}

\paragraphtitle{Dataset, Experiment Setup, and Baselines.} We evaluate on the ILSVRC 2015 VID dataset~\cite{russakovsky2015ImageNeta} using the popular YOLOv3 model~\cite{redmon2016You} with pre-trained weights from~\cite{sabater2020Robust}. We report all results on the validation set with \num{555} videos and over \num{172}k frames, using mean Average Precision (mAP) with an IoU threshold of \num{0.5} (following previous works~\cite{chen2020memory,sabater2020Robust,zhu2017Deep}). We evaluate the same model variants as in \autoref{sec:video_pose_estimation} (conventional, EvNet, Skip-Conv, and Skip-Conv reset).

\paragraphskip
\paragraphtitle{Implementation Details.} We resize all videos to $\num{224} \times \num{384}$, padding as needed to preserve the aspect ratio. Unlike OpenPose, YOLOv3 includes batch normalization (BN) layers. BN gives us a convenient way to estimate the distribution of activations at each neuron. We use this information to adjust the threshold values. Specifically, we scale the threshold at each neuron by $1 / \gamma$ (where $\gamma$ is the learned BN re-scaling parameter). This scaling makes the policy more sensitive for neurons with a narrower activation distribution, where we would expect equal-sized changes to be more meaningful.

\paragraphskip
\paragraphtitle{Results.} \autoref{fig:pareto_curves} presents our results with varying thresholds. Again, we observe that our EvNets outperform Skip-Conv variants, and sometimes have slightly higher accuracy than the conventional model with greatly reduced compute cost. \autoref{tab:accuracy_and_computation} presents the accuracy and computation at the best operating points.

\subsection{Low-Level Vision Tasks}

\begin{figure}[t]
    \centering
    \includegraphics{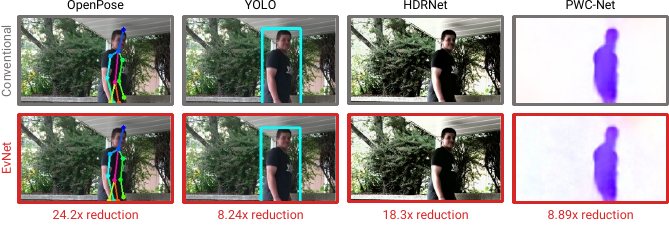}
    \caption{\textbf{Versatility of EvNets.} We demonstrate that EvNets are an effective strategy for many high- and low-level vision tasks. Across tasks, we see significant computation savings while maintaining high-quality output. This frame shows a person mid-jump. The EvNet tracks the subject correctly under rapid motion.}
    \label{fig:sample_outputs}
\end{figure}

We have so far considered only high-level inference tasks. However, EvNets are also an effective strategy for low-level vision. We consider PWC-Net~\cite{sun2018PWCNet} for optical flow computation and HDRNet~\cite{gharbi2017Deep} for video frame enhancement. For brevity, we only show sample results in \autoref{fig:sample_outputs} and refer the reader to the supplementary material for more details. As with the high-level models, we observe minimal degradation in accuracy and significant computation savings.

\subsection{Ablation and Analysis}
\label{sec:ablation_and_analysis}

\paragraphtitle{Rounding Policy Comparison.} \autoref{fig:policy_design}~(b) compares our transmission policy and the rounding policy used in a Sigma-Delta~network~\cite{oconnor2016Sigma}. We obtain these results by evaluating the fully-connected model from the Sigma-Delta paper (with the authors' original weights) on the Temporal MNIST dataset~\cite{oconnor2016Sigma}. We evaluate EvNets with thresholds of the form $10^p$, where $p \in $ \{\num{-1.5}, \num{-1.4}, $\ldots,$ \num{-0.3}, \num{-0.2}\}. We obtain results for the Sigma-Delta network using the original authors' code, which involves training the quantization threshold (the Pareto frontier is a consequence of varying a training penalty scale $\lambda$).

\begin{figure}[t]
    \centering
    \includegraphics{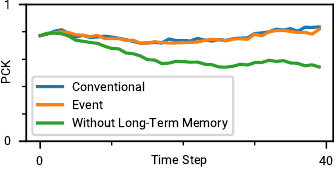}
    \caption{\textbf{Ablation of Long-Term Memory.} Removing the memory $d$, as considered in Skip-Conv~\cite{habibian2021Skipconvolutions}, causes a rapid decay in accuracy. Results using the OpenPose model on the JHMDB dataset. See \autoref{sec:experiments} for details.}
    \label{fig:accuracy_over_time}
\end{figure}

\paragraphskip
\paragraphtitle{Ablation of Long-Term Memory.} \autoref{fig:accuracy_over_time} shows the effect of ablating the long-term memory $d$ (resetting it to zero after each input). We evaluate the OpenPose model on the JHMDB dataset. Other than resetting $d$, the two models shown are identical. Both models use a threshold of \num{0.05}. We see that long-term memory is critical for maintaining stable accuracy.

\paragraphskip
\paragraphtitle{Camera Motion.} Global camera or scene motion (e.g., camera shake or scene translation) reduces the amount of visual persistence in a video. We would therefore expect camera motion to reduce the savings in an EvNet. To confirm this, we evaluate the OpenPose and YOLO models on a custom-labeled video dataset. We label the camera motion in each video as ``none'' (perfectly stationary camera), ``minor'' (slight camera shake), or ``major.'' See the supplement for details. We test OpenPose with a threshold of \num{0.05} and YOLO with a threshold of \num{0.06}. Because this dataset does not have frame-level labels for pose or object detection, we do not explicitly evaluate task accuracy. However, the thresholds we use here give good accuracy on JHMDB and VID. For OpenPose, the computation savings for ``none,'' ``major,'' and ``minor'' camera motion are \num{17.3}$\times$, \num{11.3}$\times$, and \num{8.40}$\times$, respectively. For YOLO, the savings are \num{6.64}$\times$, \num{3.95}$\times$, and \num{2.65}$\times$. As expected, we see a reduction in savings when there is strong camera motion, although we still achieve large reductions relative to the conventional model.

\paragraphskip
\paragraphtitle{Wall-Time Savings.} We now show preliminary results demonstrating wall-time savings in EvNets. We consider the HRNet~\cite{sun2019deep} model (see supplementary) on the JHDMB dataset. We evaluate on an Intel Core i7 8700K CPU.

We implement the model in PyTorch. For the EvNet, we replace the standard convolution with a custom sparse C++ convolution. Our convolution uses an input-stationary design (i.e., an outer loop over input pixels) to skip zero deltas efficiently. In the conventional model, we use a custom C++ convolution with a standard output-stationary design (i.e., an outer loop over output pixels). We use a custom operator in the conventional model to ensure a fair comparison, given the substantial engineering effort invested in the default MKL-DNN library. We implement both operators with standard best practices (e.g., maximizing data-access locality). We compile with GCC~9.4 with the \texttt{-Ofast} flag.

For evaluation, we use an input size of $\num{256} \times \num{256}$ and an EvNet threshold of \num{0.1}. The EvNet achieves a PCK of \SI{90.46}{\percent} and runs in an average of \SI{0.3497}{\second} (\num{7.361e8}~ops) per frame. The conventional model achieves a PCK of \SI{90.37}{\percent} and runs in \SI{1.952}{\second} (\num{1.019e10}~ops) per frame.

\section{Discussion}
\label{sec:discussion}

\paragraphtitle{Hardware Platforms.} Mainstream GPU hardware is designed for parallel, block-wise computation with coarse control flow. EvNets with neuron-level transmission are inefficient under this computation model. In the long term, we expect to achieve the best performance on specialized hardware designed for extreme parallelism and distributed control. It is important to emphasize that event neurons \emph{do not need to operate by a shared clock}. Each neuron operates independently -- consuming new input as it arrives and transmitting output once it is computed. This independence permits an asynchronous, networked execution model in contrast to the ordered, frame-based model in conventional machine learning. Spiking neural networks (SNNs)~\cite{maass1997Networks} share this asynchronous computation model and have motivated the development of neuromorphic hardware platforms \cite{akopyan2015TrueNorth,davies2018Loihi} that could be re-purposed for efficient implementation of EvNets. 

\paragraphskip
\paragraphtitle{Acknowledgments.} This research was supported by NSF CAREER Award 1943149.

\bibliographystyle{splncs04}
\bibliography{references}

\end{document}


\maketitle

This is the supplement to our main paper. Here we present further results on low-level vision tasks (\autoref{sec:results_on_low-level_tasks}), show additional analysis experiments (\autoref{sec:additional_analysis_experiments}), show preliminary results on HRNet~\cite{sun2019deep}, provide several example model outputs (\autoref{sec:example_outputs}), expand on the details of our main experiments (\autoref{sec:experiment_details}), provide a derivation for Eq.~4 (\autoref{sec:derivation_of_equation_4}), and discuss the theoretical properties of event networks (\autoref{sec:thoughts_on_theoretical_guarantees}). For sections, figures, tables, and equations, we use numbers (e.g., Fig.~1) to refer to the main paper and capital letters (e.g., Fig.~A) to refer to this supplement.

\section{Results on Low-Level Tasks}
\label{sec:results_on_low-level_tasks}

In this section, we describe our experiments for low-level vision tasks. We consider HDRNet for image enhancement \cite{gharbi2017Deep} and PWC-Net for optical flow \cite{sun2018PWCNet}.

Note that these models include some specialized operations (i.e., the bilateral transform for HDRNet~\cite{gharbi2017Deep} and flow warping in PWC-Net~\cite{sun2018PWCNet}). These operations represent a small portion of the overall computational cost of the models. For simplicity, we exclude them when counting multiply-accumulate operations.

\paragraphskip
\paragraphtitle{Image Enhancement.} HDRNet~\cite{gharbi2017Deep} can be trained to reproduce several image enhancement effects. We use the Local~Laplacian~\cite{paris2011Local} version of the model. HDRNet has two subnetworks: a deep, low-resolution feature network and a shallow, high-resolution guidemap network. The guidemap network represents only about \SI{10}{\percent} of the overall operations, and converting it to an EvNet has a noticeable effect on the visual quality of the output. Therefore, we only convert the feature network to an EvNet. We report operation savings for both the overall model (both subnetworks) and the feature network (the EvNet portion). We refer to these operation counts as ``HDRNet-a'' and ``HDRNet-f,'' respectively. We use a threshold of $h = 0.1$ and evaluate using the PSNR metric. We resize all images to $540 \times 960$ before applying the model.

We use the original authors' pretrained weights. However, these weights were trained on a non-public dataset. Therefore, instead of evaluating the model against ground truth labels, we compute the agreement between the outputs of the event model and conventional model. We evaluate on a subset of the MPII video dataset~\cite{andriluka20142D} (see \autoref{sec:experiment_details} for details on the dataset). See \autoref{tab:results_on_low-level_tasks} and \autoref{tab:camera_motion_for_low-level_tasks} for results. We also show example model outputs \autoref{fig:hdrnet_result_samples}.

\paragraphskip
\paragraphtitle{Optical Flow.} We also consider the PWC-Net~model~\cite{sun2018PWCNet} for optical flow computation.  Unlike the other models (OpenPose, YOLO, HDRNet) which take a single frame as input, this model takes a \emph{pair} of frames. We use a threshold of $h = 0.01$ and evaluate using the EPE~metric~\cite{baker2011Database}. We resize all images to $288 \times 512$ before applying the model. We use the original authors' weights trained on Sintel~\cite{butler2012Naturalistic}. Like with HDRNet, we evaluate the agreement between the event and conventional outputs. Results are shown in \autoref{tab:results_on_low-level_tasks} and \autoref{tab:camera_motion_for_low-level_tasks}. We also evaluate on the ground-truth labels in the Sintel training dataset. On this data, the conventional model achieves EPE \num{2.86} and the event model achieves EPE \num{3.33}.

\begin{table}[t]
    \centering
    \caption{\textbf{Results on Low-Level Tasks.} Results for image enhancement and optical flow. The tables gives the overall computation savings, agreement between the conventional and event models, and overhead percentages (number of extra operations expended for each operation saved).}
    \label{tab:results_on_low-level_tasks}
    \begin{tabular}{lcccc}
        \toprule
        Model    & Savings     & Agreement       & Math                & Load/Store          \\ \midrule
        HDRNet-a & \num{5.78}x & \num{39.4} PSNR & \SI{2.57}{\percent} & \SI{3.79}{\percent} \\
        HDRNet-f & \num{23.9}x & \num{39.4} PSNR & \SI{2.57}{\percent} & \SI{3.79}{\percent} \\
        PWC-Net  & \num{2.68}x & \num{0.335} EPE & \SI{0.44}{\percent} & \SI{0.74}{\percent} \\
        \bottomrule
    \end{tabular}
\end{table}

\begin{table}[t]
    \centering
    \caption{\textbf{Camera Motion for Low-Level Tasks.} The savings factor for different levels of camera motion, evaluated on our custom MPII dataset (see \autoref{sec:experiment_details}).}
    \label{tab:camera_motion_for_low-level_tasks}
    \begin{tabular}{lccc}
        \toprule
        Model    & None               & Minor              & Major              \\ \midrule
        HDRNet-a & \num{6.19}$\times$ & \num{6.02}$\times$ & \num{5.55}$\times$ \\
        HDRNet-f & \num{34.9}$\times$ & \num{29.7}$\times$ & \num{20.0}$\times$ \\
        PWC-Net  & \num{5.41}$\times$ & \num{3.29}$\times$ & \num{2.11}$\times$ \\
        \bottomrule
    \end{tabular}
\end{table}

\section{Additional Analysis Experiments}
\label{sec:additional_analysis_experiments}

\begin{figure}
    \centering
    \includegraphics{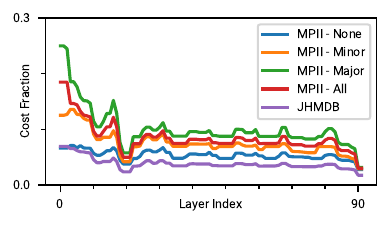}
    \caption{\textbf{Operation Costs by Layer.} Results for the OpenPose model on the JHMDB and custom-labelled MPII datasets. The increasing savings with depth show the importance of leveraging repetition at all levels of the network hierarchy. We have applied a median filter of size \num{5} (along the layer axis) to the data in this plot.}
    \label{fig:openpose_ops_layer}
\end{figure}

\paragraphtitle{Layer Trends.} \autoref{fig:openpose_ops_layer} shows the computational cost of the OpenPose model as a function of the layer depth. We show results both on the JHMDB dataset and on our custom-labelled MPII dataset (to allow analysis of the effect of camera motion). Overall, we see a reduction in the relative cost as we go deeper in the network. This highlights the importance of leveraging repetition in the deep layers of the network, not just near the input. We also observe that the early layers transmit more frequently when there is large camera motion. This corresponds to an increased number of changes in low-level features and pixel values.

\begin{figure}
    \centering
    \includegraphics{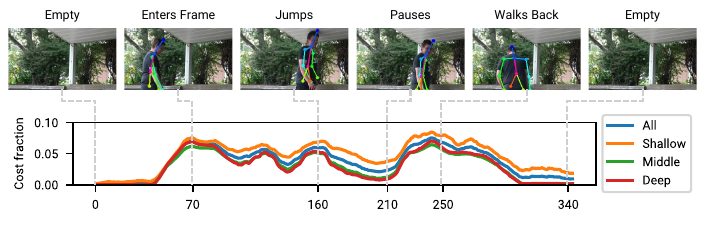}
    \caption{\textbf{Temporal Variation in Operation Cost.} Identifiable events in the video (e.g., jumping) correspond to temporary increases in the number of operations. ``Shallow'' corresponds to the first \num{31} layers, ``middle'' to the next \num{31}, and ``deep'' to the final \num{30}. We have applied a centered moving average of size \num{10} (along the time axis) to the data in this plot.}
    \label{fig:temporal_variation}
\end{figure}

\paragraphskip
\paragraphtitle{Temporal Variation.} \autoref{fig:temporal_variation} shows the per-frame computational cost of the OpenPose EvNet over the course of a video. The video in question has a static background and a moving foreground object (person). Recognizable events in the video (e.g., walking, jumping) correspond to temporary increases in the number of operations. In this way, we see EvNets living up to their promise of ``only computing when something interesting is happening.''

\paragraphskip
\paragraphtitle{Varying Granularity.} \autoref{tab:varying_granularity} shows the effect of increasing the granularity of the policy. We evaluate the OpenPose~model~\cite{cao2017Realtime} on the JHMDB dataset~\cite{jhuang2013Understanding}. We test both a spatial chunking policy and a policy that chunks along the channel dimension (e.g., \cite{habibian2021Skipconvolutions}). Because each neighborhood computes a mean of several $|d|$, the thresholds must be reduced to keep the accuracy from dropping. The threshold-setting strategy $0.05 / \sqrt{n}$ is a heuristic that we found to give relatively stable accuracy with varying $n$. The results show that increasing the chunk size reduces the operation savings. However, chunking may, in practice, allow more efficient execution on certain hardware.

\begin{table}[t]
    \centering
    \caption{\textbf{Varying Granularity.} Results for the OpenPose model on the JHMDB dataset. Using larger chunks, especially channel chunks, reduces the computational gains somewhat. However, chunking may have practical benefits on some hardware.}
    \label{tab:varying_granularity}
    \begin{tabular}{lccc}
        \toprule
        Variant        & Threshold         & PCK          & Operations     \\ \midrule
        Conventional   & --                & \num{0.7640} & \num{7.055e10} \\
        No chunking    & $0.05$            & \num{0.7581} & \num{6.166e9}  \\
        2x2 chunks     & $0.05 / \sqrt{2}$ & \num{0.7575} & \num{8.574e9}  \\
        4x4 chunks     & $0.05 / \sqrt{4}$ & \num{0.7662} & \num{1.191e10} \\
        8x8 chunks     & $0.05 / \sqrt{8}$ & \num{0.7431} & \num{1.646e10} \\
        Channel chunks & $0.02$            & \num{0.7600} & \num{1.782e10} \\
        \bottomrule
    \end{tabular}
\end{table}

\paragraphskip
\paragraphtitle{Comparison Against Output Interpolation.} One alternate strategy for efficient video inference is to run a model once every $n$ frames and interpolate its predictions for the remaining $n - 1$ frames. We apply this strategy to OpenPose on JHMDB and compare it to the EvNet approach. We use $n = 16$ and linearly interpolate the joint positions between model predictions (the value \num{16} was chosen to give a computational cost close to the EvNet in Table~1). The interpolated model expends \num{6.764e9}~ops per frame on average and achieves a PCK of \SI{68.52}{\percent} (a reduction of \SI{7.88}{\percent} from the conventional model). Compare this to the EvNet in Table~1, which expends \num{6.780e9}~ops on average while achieving a PCK score of \SI{76.37}{\percent} (a reduction of \SI{0.03}{\percent} from the conventional model). Compared to output interpolation, the EvNet gives much higher accuracy at a similar computation cost. Note that we trim the inputs for the interpolation model to have a length of $kn + 1$ frames, where $k$ is a positive integer. This ensures that the video can be divided into uniform blocks of $n$ frames (with one extra frame at the end). If we trim to the same length for the EvNet, it achieves a PCK of \SI{76.82}{\percent} with average cost \num{7.265e9}~ops. The conventional model achieves PCK \SI{76.67}{\percent} at \num{7.055e10}~ops on the trimmed video.

\paragraphskip
\paragraphtitle{Temporal Smoothness.} We have anecdotally observed improved temporal smoothness in the outputs of EvNets. We hypothesize that this is one of the reasons for the slightly increased accuracy for some models (e.g., Table~1) over the conventional baselines. We quantitatively measure temporal smoothness for OpenPose on JHMDB by measuring the mean L2 joint motion between frames. The average joint motion for the conventional model is \num{10.3} pixels. For the EvNet with threshold $h = $ \{\num{0.01}, \num{0.02}, \num{0.04}, \num{0.06}, \num{0.08}\}, the average motion was \{\num{9.77}, \num{9.26}, \num{7.98}, \num{7.14}, \num{5.81}\} pixels. This confirms that the EvNet outputs are more temporally smooth than those of the conventional model, with smoothness increasing with the policy threshold.

\section{HRNet Experiments}
\label{sec:hrnet_experiments}

We test HRNet~\cite{sun2019deep}, a state-of-the-art model for various location-based tasks (e.g., object detection) on the JHMDB pose recognition dataset~\cite{jhuang2013Understanding}. We use the HRNet-W32 version of the model.

\paragraphskip
\paragraphtitle{Training Procedure.} We initialize with pretrained MPII weights from~\cite{sun2019deep}. We fine-tune the model on JHMDB for \num{20} epochs using the Adam optimizer and a learning rate of \num{1e-5}. We set aside \SI{20}{\percent} of the training data for validation and save the model at the epoch with the lowest validation loss. JHMDB defines three train/test splits -- we train and evaluate a model on each training split and average the results (accuracy and computation costs) over the three splits. Where not otherwise specified, we adopt the training and data augmentation parameters of~\cite{habibian2021Skipconvolutions}. All of our training code will be publicly released and included with the supplementary material.

\paragraphskip
\paragraphtitle{Evaluation.} We evaluate three model variants: the conventional model, an EvNet and Skip-Conv (without periodic resets). See \autoref{tab:hrnet_results} for results. We report the PCK metric (as in our experiments on OpenPose). The accuracy and savings we observe are in line with our other experiments.

\begin{table}[t]
    \centering
    \caption{\textbf{HRNet Results.} Results for the HRNet model on JHMDB.}
    \label{tab:hrnet_results}
    \begin{tabular}{lccc}
        \toprule
        Model        & Threshold  & PCK                  & Operations     \\ \midrule
        Conventional & --         & \SI{90.37}{\percent} & \num{1.019e10} \\
        EvNet        & \num{0.05} & \SI{90.43}{\percent} & \num{1.112e9}  \\
        EvNet        & \num{0.1}  & \SI{90.46}{\percent} & \num{7.361e8}  \\
        EvNet        & \num{0.2}  & \SI{86.44}{\percent} & \num{4.187e8}  \\
        Skip-Conv    & \num{0.05} & \SI{89.17}{\percent} & \num{1.035e9}  \\
        Skip-Conv    & \num{0.1}  & \SI{84.45}{\percent} & \num{6.473e8}  \\
        Skip-Conv    & \num{0.2}  & \SI{78.72}{\percent} & \num{3.307e8}  \\
        \bottomrule
    \end{tabular}
\end{table}

\section{Example Outputs}
\label{sec:example_outputs}

\begin{figure}
    \centering
    \includegraphics[width=\textwidth]{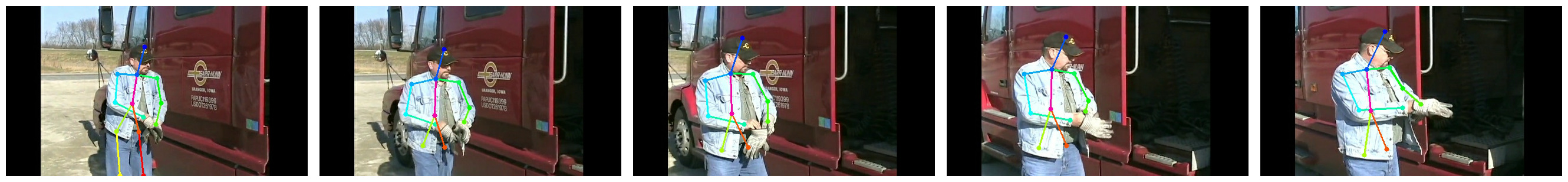} \\[-0.02in]
    \includegraphics[width=\textwidth]{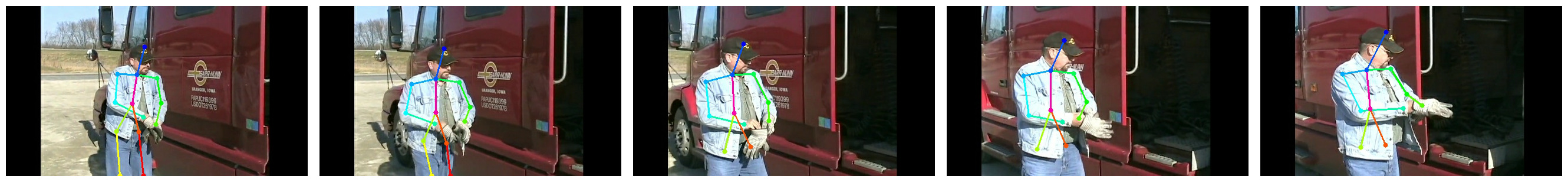} \\[-0.02in]
    \vspace*{0.1in}
    \includegraphics[width=\textwidth]{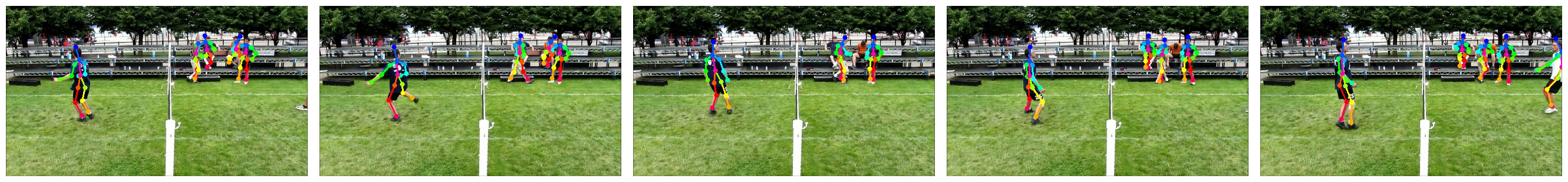} \\[-0.02in]
    \includegraphics[width=\textwidth]{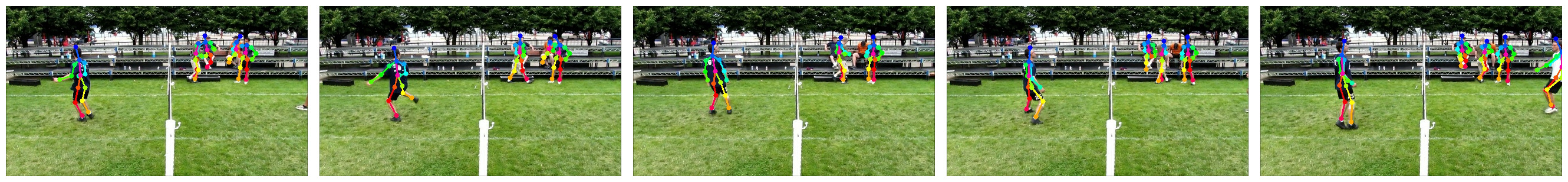} \\[-0.02in]
    \vspace*{0.1in}
    \includegraphics[width=\textwidth]{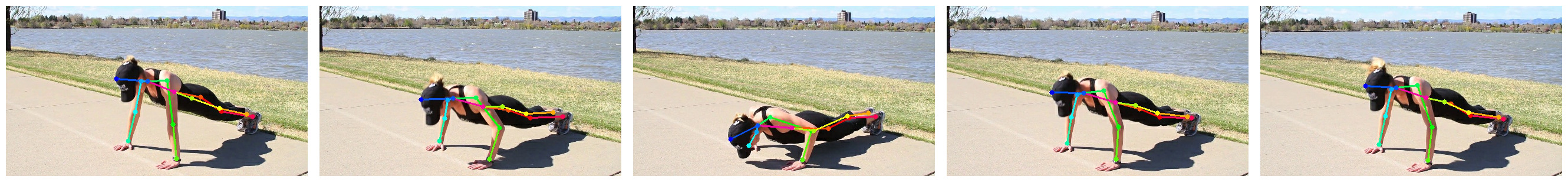} \\[-0.02in]
    \includegraphics[width=\textwidth]{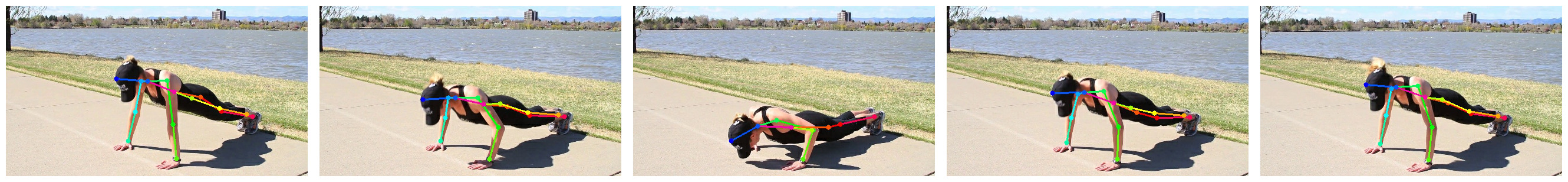} \\[-0.02in]
    \vspace*{0.1in}
    \includegraphics[width=\textwidth]{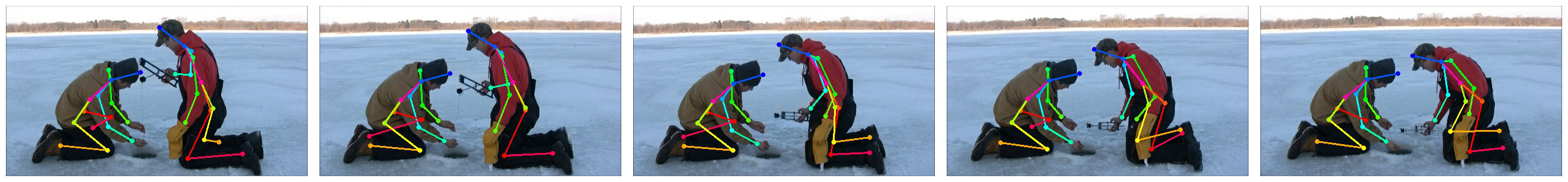} \\[-0.02in]
    \includegraphics[width=\textwidth]{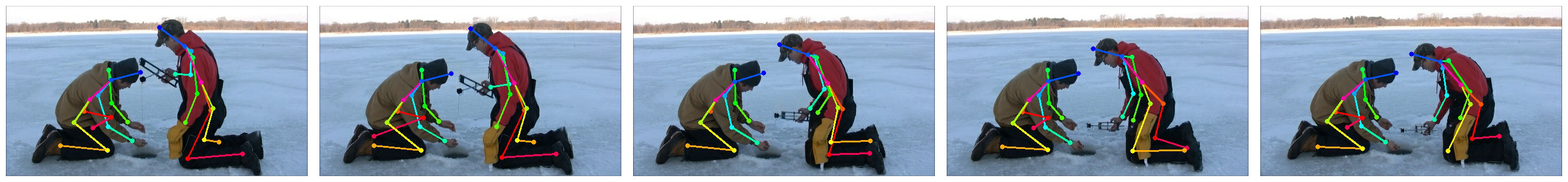} \\[-0.02in]
    \vspace*{0.1in}
    \includegraphics[width=\textwidth]{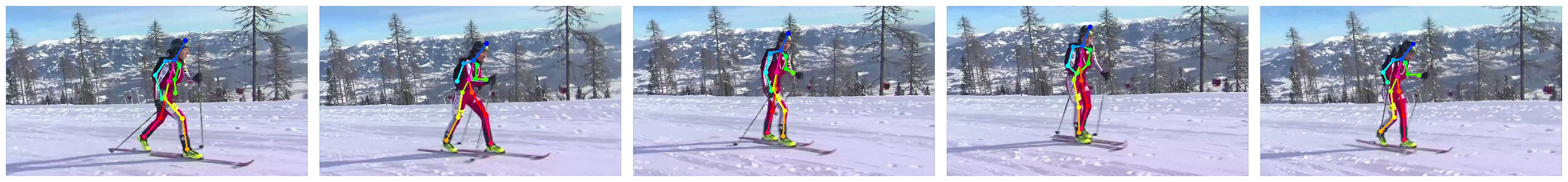} \\[-0.02in]
    \includegraphics[width=\textwidth]{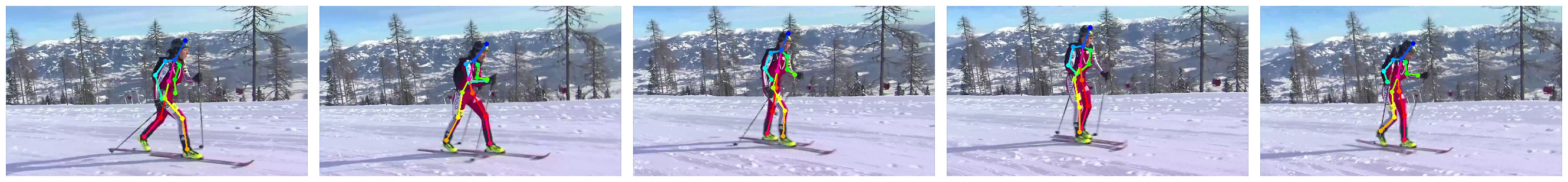} \\[-0.02in]
    \caption{\textbf{OpenPose Result Samples.} Conventional output is on the top and event output is on the bottom. We show frames \num{0}, \num{10}, \num{20}, \num{30}, and \num{40} from each video. Videos are from the MPII dataset.}
    \label{fig:openpose_result_samples}
\end{figure}

\begin{figure}
    \centering
    \includegraphics[width=\textwidth]{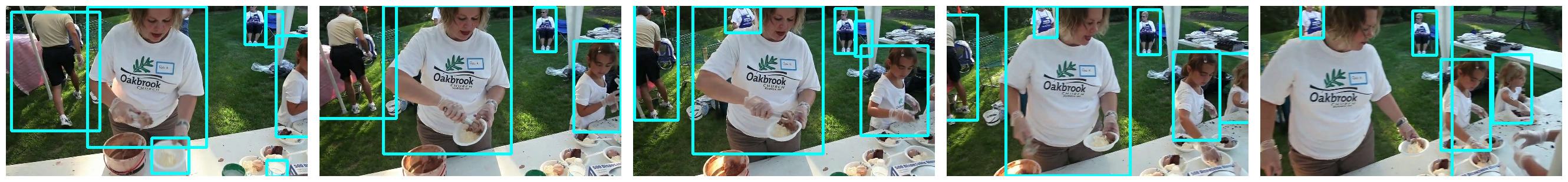} \\[-0.02in]
    \includegraphics[width=\textwidth]{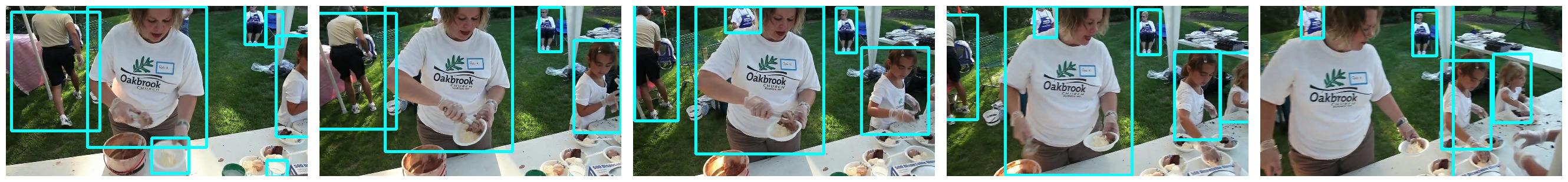} \\[-0.02in]
    \vspace*{0.1in}
    \includegraphics[width=\textwidth]{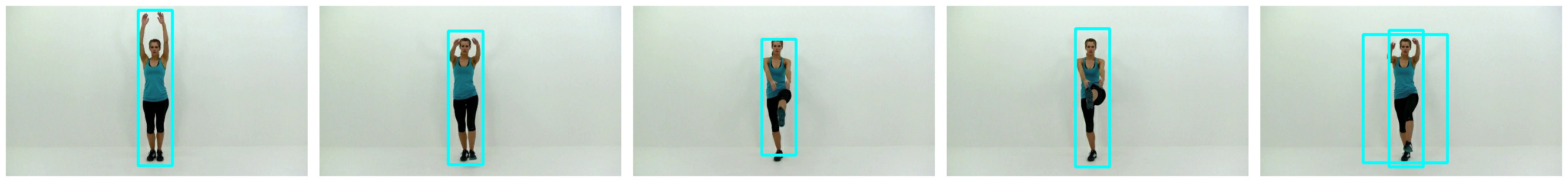} \\[-0.02in]
    \includegraphics[width=\textwidth]{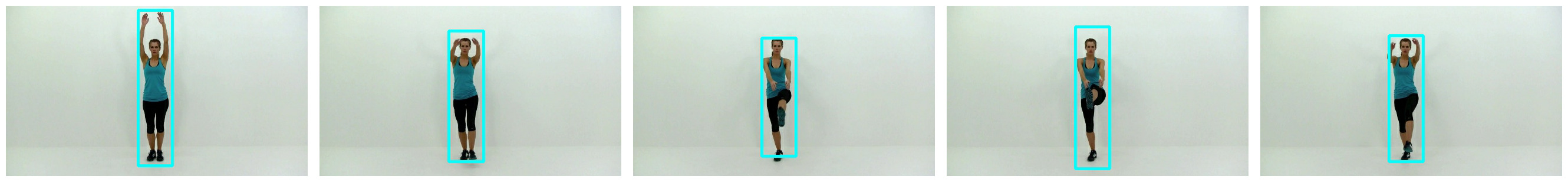} \\[-0.02in]
    \vspace*{0.1in}
    \includegraphics[width=\textwidth]{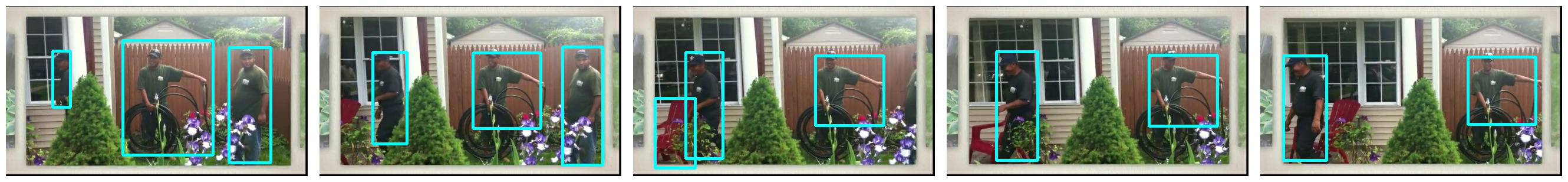} \\[-0.02in]
    \includegraphics[width=\textwidth]{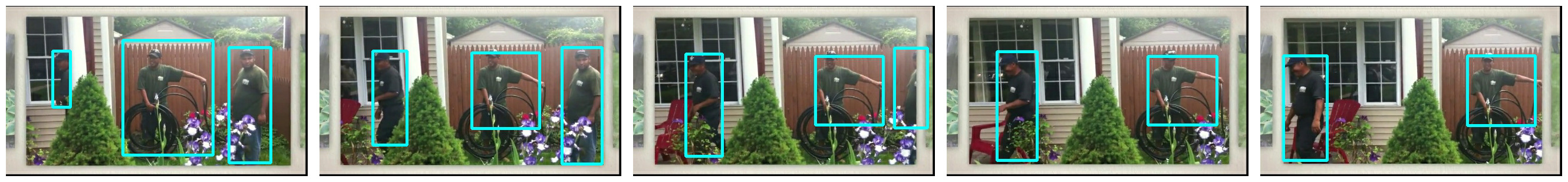} \\[-0.02in]
    \vspace*{0.1in}
    \includegraphics[width=\textwidth]{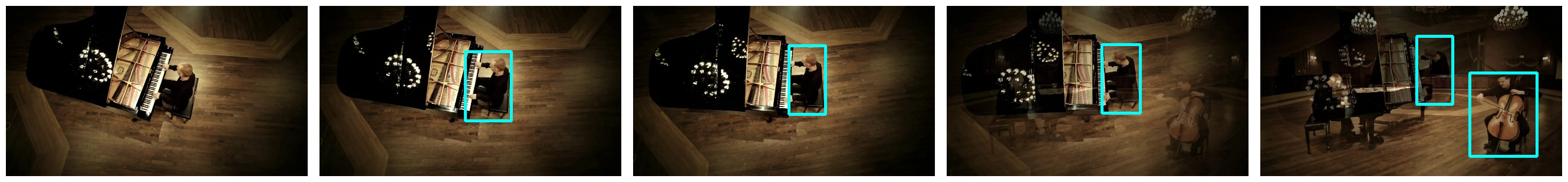} \\[-0.02in]
    \includegraphics[width=\textwidth]{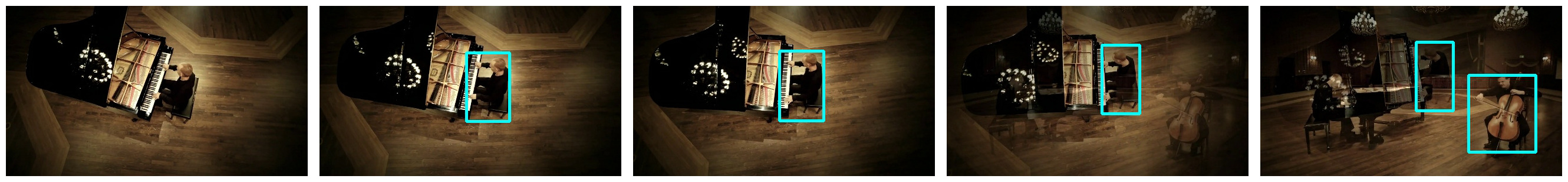} \\[-0.02in]
    \vspace*{0.1in}
    \includegraphics[width=\textwidth]{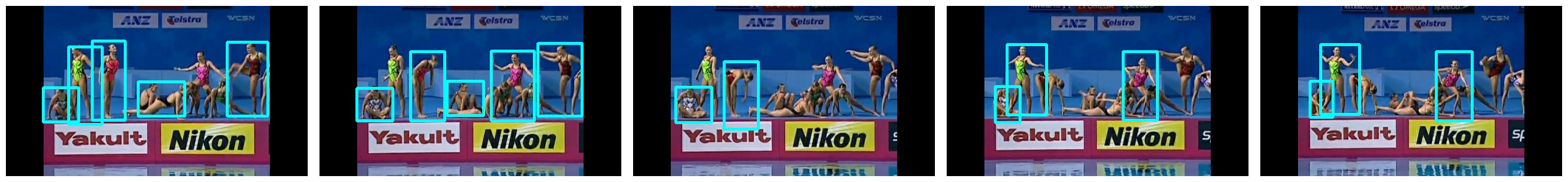} \\[-0.02in]
    \includegraphics[width=\textwidth]{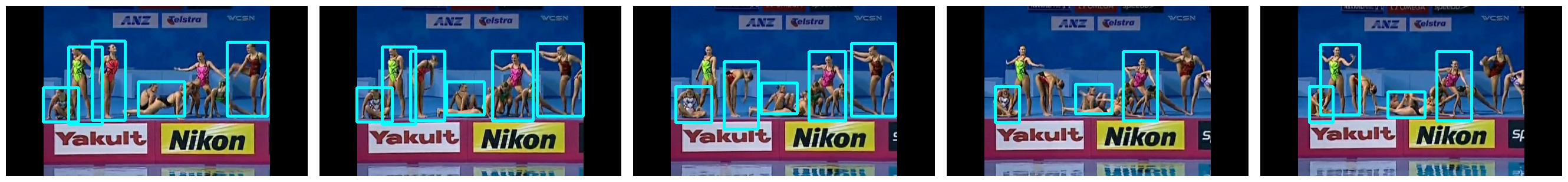} \\[-0.02in]
    \caption{\textbf{YOLO Result Samples.} Conventional output is on the top and event output is on the bottom. We show frames \num{0}, \num{10}, \num{20}, \num{30}, and \num{40} from each video. Videos are from the MPII dataset.}
    \label{fig:yolo_result_samples}
\end{figure}

\begin{figure}
    \centering
    \includegraphics[width=\textwidth]{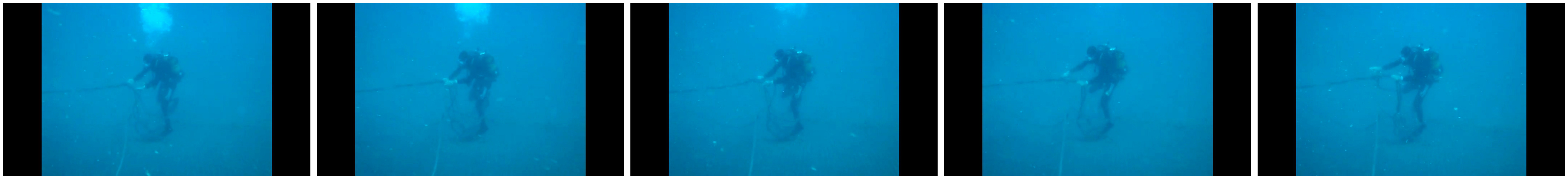} \\[-0.02in]
    \includegraphics[width=\textwidth]{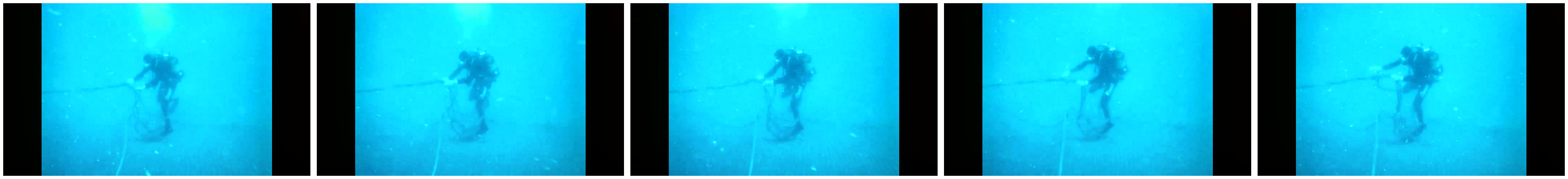} \\[-0.02in]
    \includegraphics[width=\textwidth]{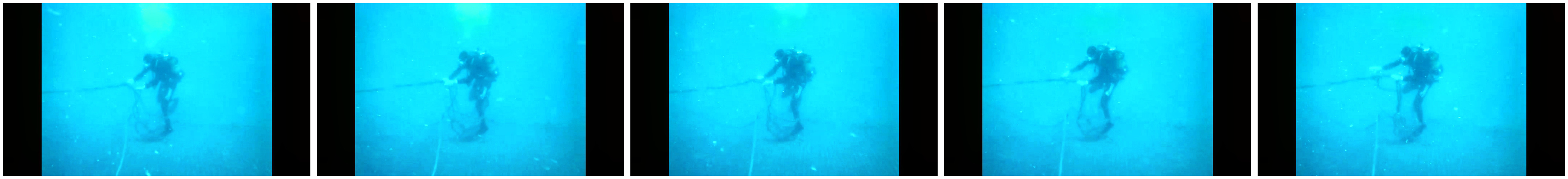} \\[-0.02in]
    \vspace*{0.1in}
    \includegraphics[width=\textwidth]{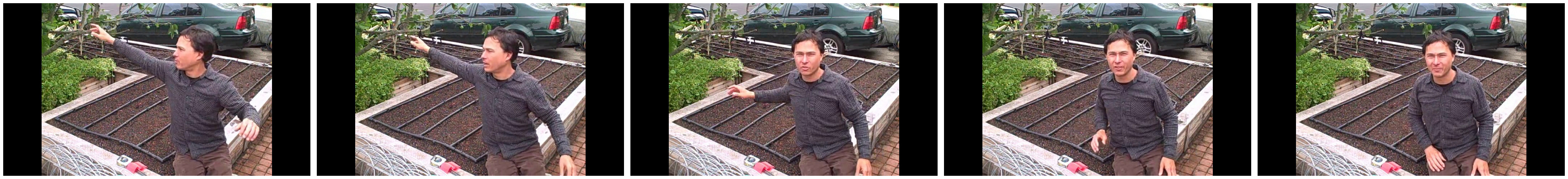} \\[-0.02in]
    \includegraphics[width=\textwidth]{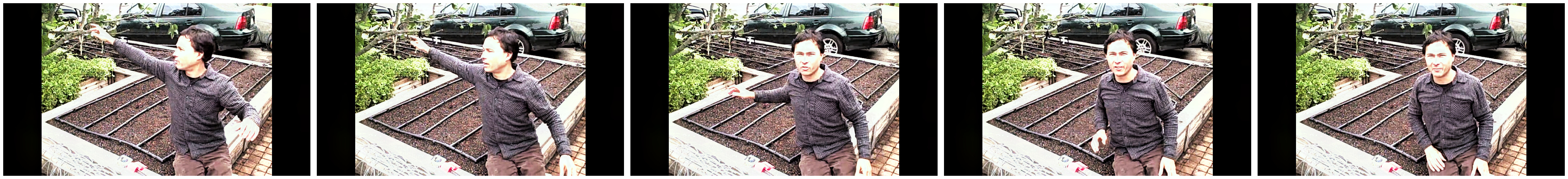} \\[-0.02in]
    \includegraphics[width=\textwidth]{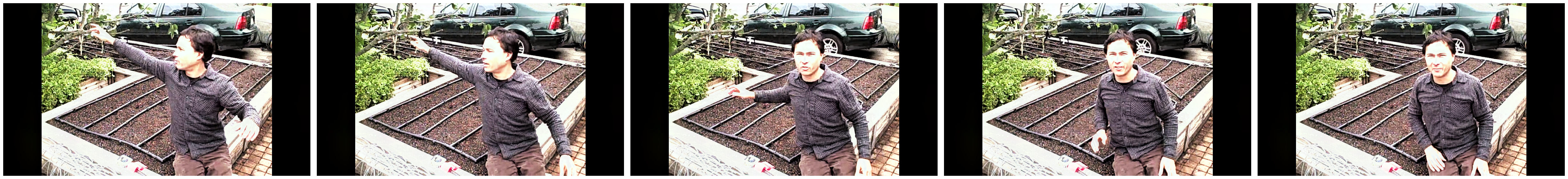} \\[-0.02in]
    \vspace*{0.1in}
    \includegraphics[width=\textwidth]{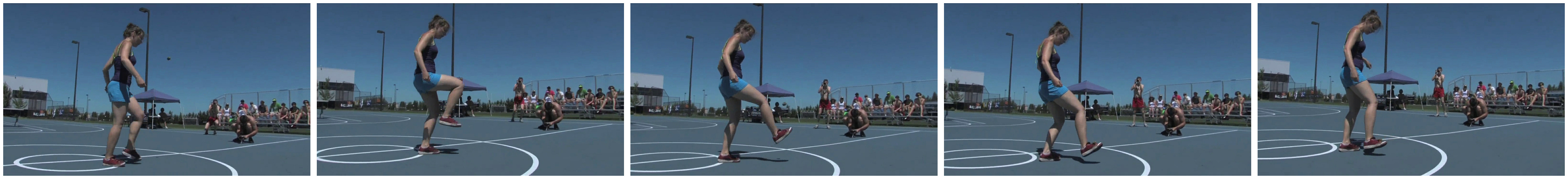} \\[-0.02in]
    \includegraphics[width=\textwidth]{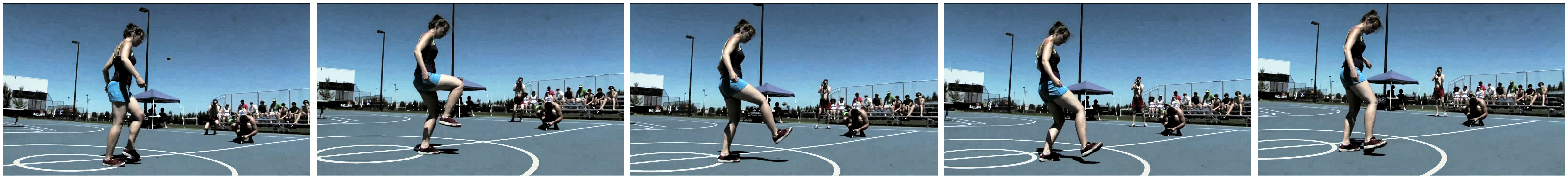} \\[-0.02in]
    \includegraphics[width=\textwidth]{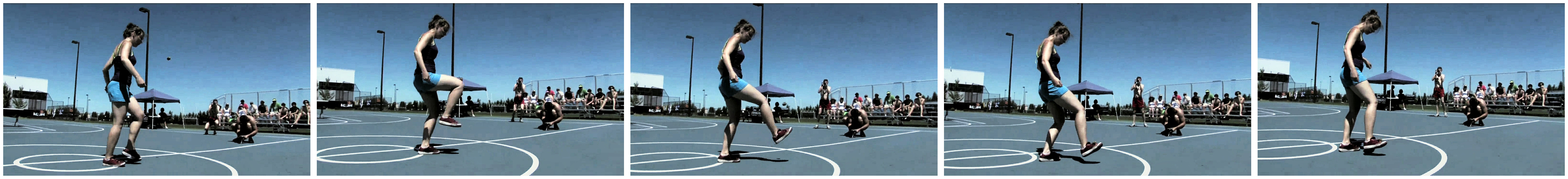} \\[-0.02in]
    \caption{\textbf{HDRNet Result Samples.} The model input is on the top, conventional output is in the middle, and event output is on the bottom. We show frames \num{0}, \num{10}, \num{20}, \num{30}, and \num{40} from each video. Videos are from the MPII dataset.}
    \label{fig:hdrnet_result_samples}
\end{figure}

\begin{figure}
    \centering
    \includegraphics[width=0.8\textwidth]{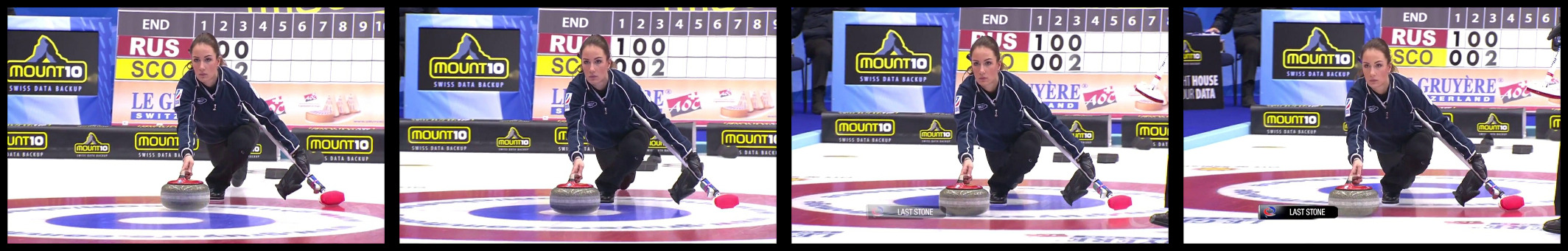} \\[-0.02in]
    \includegraphics[width=0.8\textwidth]{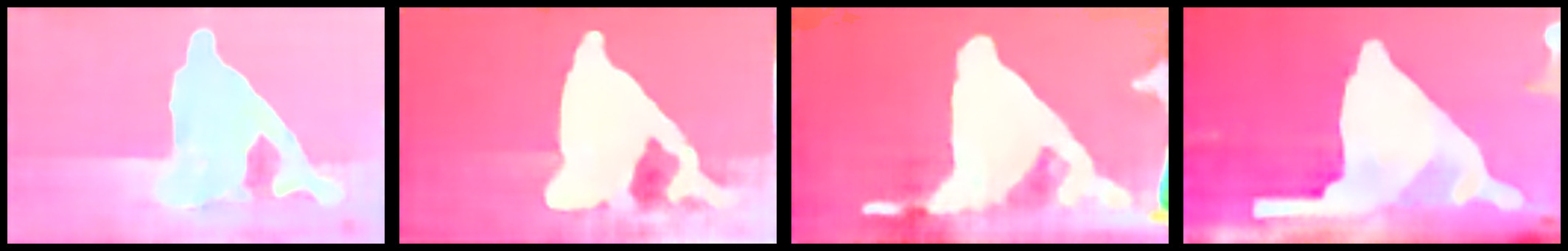} \\[-0.02in]
    \includegraphics[width=0.8\textwidth]{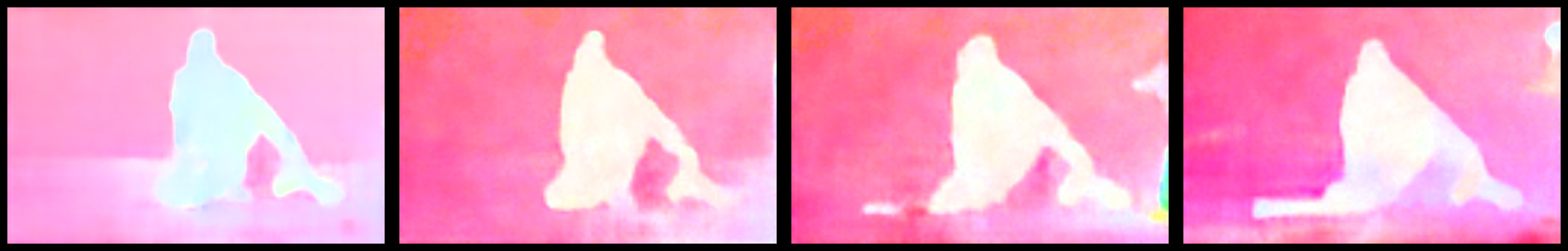} \\[-0.02in]
    \vspace*{0.1in}
    \includegraphics[width=0.8\textwidth]{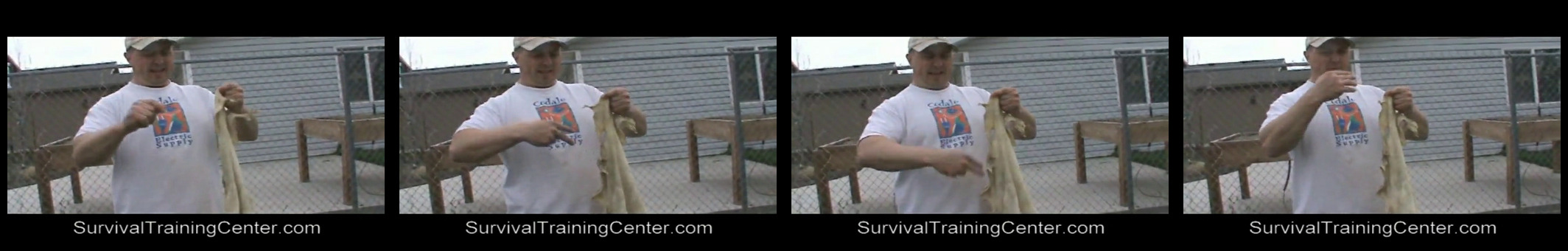} \\[-0.02in]
    \includegraphics[width=0.8\textwidth]{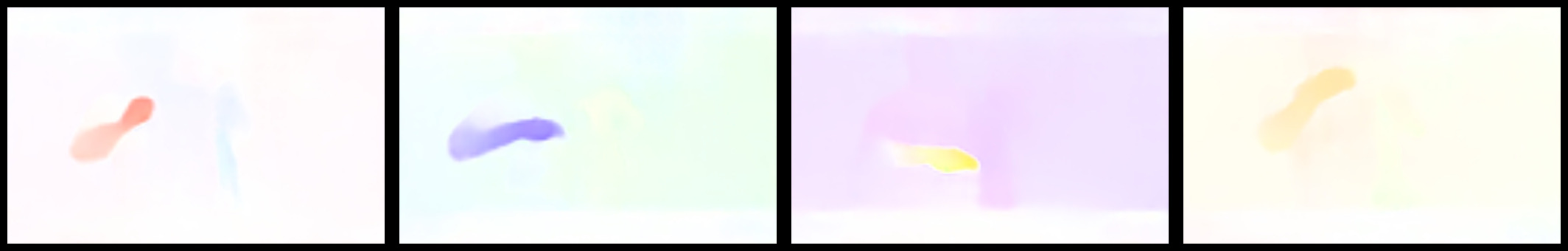} \\[-0.02in]
    \includegraphics[width=0.8\textwidth]{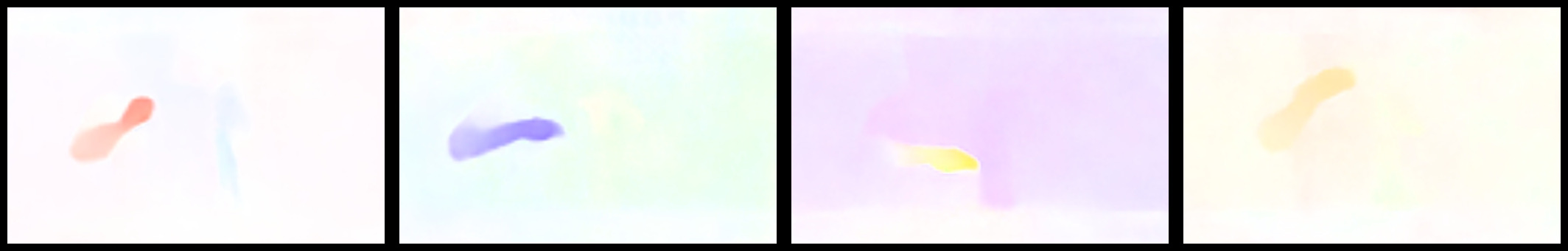} \\[-0.02in]
    \vspace*{0.1in}
    \includegraphics[width=0.8\textwidth]{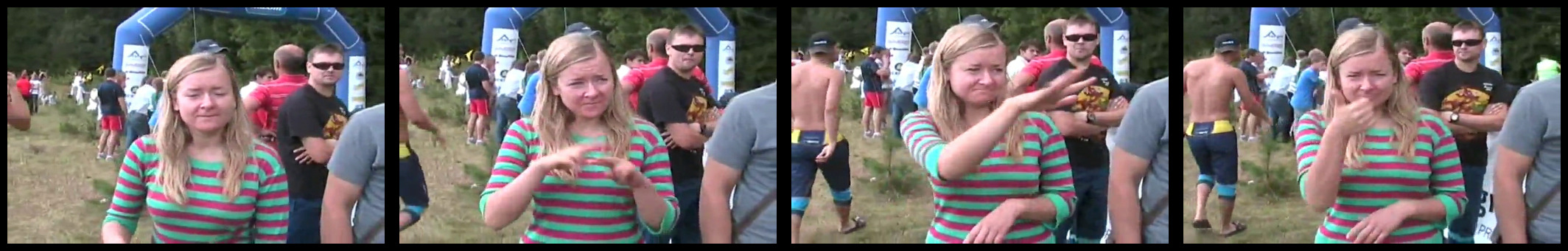} \\[-0.02in]
    \includegraphics[width=0.8\textwidth]{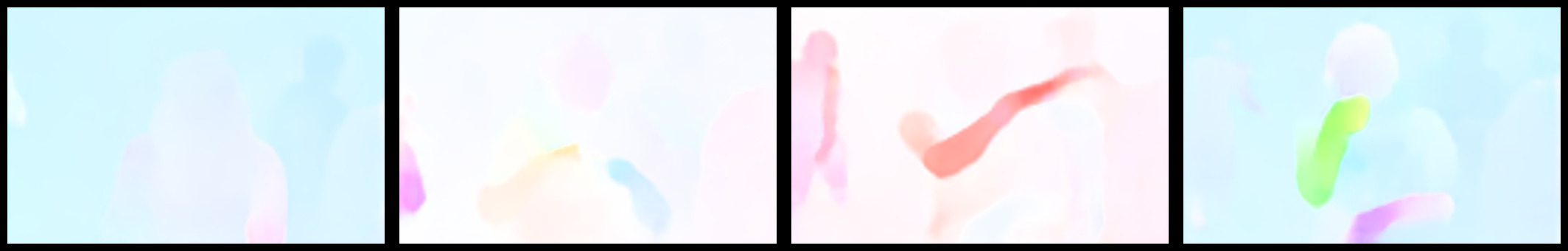} \\[-0.02in]
    \includegraphics[width=0.8\textwidth]{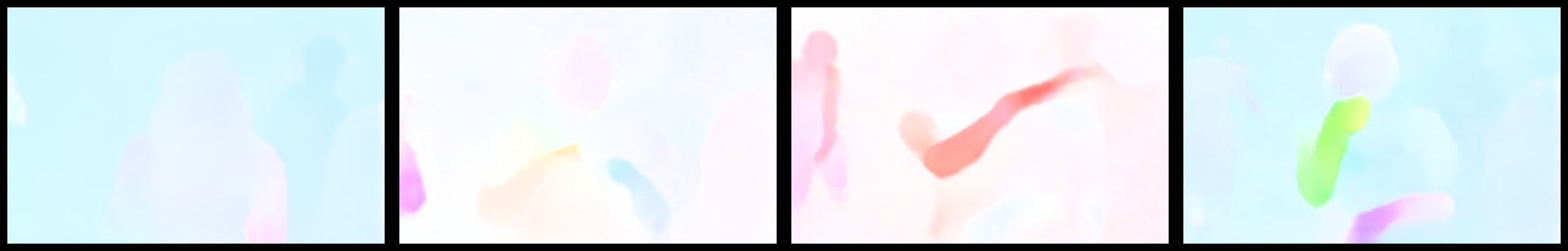} \\[-0.02in]
    \caption{\textbf{PWC-Net Result Samples.} The model input is on the top, conventional output is in the middle, and event output is on the bottom. We show frames \num{0}, \num{10}, \num{20}, and \num{30} from each video. Videos in this dataset have \num{41} frames, and we predict flow for each pair of frames. There, we cannot show output for frame \num{40}. Videos are from the MPII dataset.}
    \label{fig:pwc-net_result_samples}
\end{figure}

\paragraphskip
\paragraphtitle{Example Outputs.} \autoref{fig:openpose_result_samples}, \autoref{fig:yolo_result_samples}, \autoref{fig:hdrnet_result_samples}, and \autoref{fig:pwc-net_result_samples} show several example outputs for OpenPose, YOLO, HDRNet, and PWC-Net, respectively. The videos shown are from our custom MPII dataset (and hence all have \num{41} frames). In general, we see strong agreement between the conventional and event predictions. In some cases (especially with the YOLO model; \autoref{fig:yolo_result_samples}), we observe greater consistency in the EvNet predictions across frames. This is a consequence of an event network's preference for re-using previous activation values. This greater temporal consistency does not appear to reduce the model's ability to keep up with rapid changes.

\section{Experiment Details}
\label{sec:experiment_details}

\paragraphskip
\paragraphtitle{Custom MPII Dataset.} Here we describe the dataset that we use in our camera motion experiments (Table~3 and \autoref{tab:camera_motion_for_low-level_tasks}) and for evaluating HDRNet and PWC-Net (\autoref{tab:results_on_low-level_tasks}). We take a subset of the MPII~video~dataset~\cite{andriluka20142D} -- specifically, the first \num{246} videos that have exactly \num{41} frames (most, but not all videos in MPII have \num{41} frames). We then label each video in this dataset as having ``no camera motion'' (perfectly stationary camera), ``minor camera motion'' (slight camera shake), or ``major camera motion''. These splits contain \num{59}, \num{46}, and \num{141} videos, respectively.

\paragraphskip
\paragraphtitle{Overhead Counting.} We count overhead operations as follows. An update to an accumulator requires one load ($a$), one addition ($a + g(\bm{\Delta}_\text{in})$), and one store ($a$). An update to a gate requires two loads ($b$ and $d$), three additions ($d + f(a) - b$ and $|d| - h$), and two stores ($b$ and $d$). A transmission requires one load ($d$), one subtraction ($d - \Delta_\text{out}$), and one store ($d$).

\paragraphskip
\paragraphtitle{Tables of Results.} \autoref{tab:video_pose_estimation} shows the complete results for OpenPose on JHMDB. These values correspond to the points in Fig.~7~(a). \autoref{tab:video_pose_estimation} shows the complete results for YOLO~\cite{redmon2016You} on VID~\cite{russakovsky2015ImageNeta}, corresponding to Fig.~7~(b). \autoref{tab:operation_overhead} shows the overhead operation percentages for all thresholds tested for Fig.~7.

We also show results for larger input images ($\num{352} \times \num{480}$ for OpenPose and $\num{320} \times \num{544}$ for YOLO). Results for pose recognition, object detection, and overhead are given in \autoref{tab:video_pose_estimation_for_larger_images}, \autoref{tab:video_object_detection_for_larger_images}, and \autoref{tab:operation_overhead_for_larger_images}, respectively.

\section{Derivation of Equation 4}
\label{sec:derivation_of_equation_4}

The equation for $a^{(T)}$ is a consequence of the update to $a^{(t)}$ defined in Eq.~3, combined with the linearity of $g$ ($g$ of a sum is equal to the sum of the $g$). The equation for $b^{(T)}$ is a direct consequence of the update rule in Eq.~3.

The equation for $d^{(T)}$ in Eq.~4 comes from combining Eq.~3 and the post-transmission subtraction of $\Delta_\text{out}$. Let $d^{(0)} = 0$ as stated in Sec.~4.2. With Eq.~3, and noting that $b^{(t)} = f(a^{(t)})$,
\begin{equation}
    \textstyle d^{(T)} = \sum_{t = 1}^T \left( f(a^{(t)}) - b^{(t - 1)} \right) = b^{(T)} - b^{(0)}.
\end{equation}
Adding in the post-transmission subtraction of $\Delta_\text{out}^{(t)}$, we have
\begin{equation}
    \textstyle d^{(T)} = b^{(T)} - b^{(0)} - \sum_{t = 1}^T \Delta_\text{out}^{(t)}.
\end{equation}

\section{Thoughts on Theoretical Guarantees}
\label{sec:thoughts_on_theoretical_guarantees}

For certain special cases of transmission policies (e.g., a threshold policy with $h = 0$), we can guarantee that the output of an EvNet will be equal to that of the equivalent conventional network. As we make the policy more selective (e.g., by increasing $h$), the efficiency of the EvNet improves, but its output increasingly deviates from that of the conventional network. While we currently describe this behavior qualitatively, developing the rigorous theoretical tools necessary for a quantitative description is an important next step.

We can describe a neural network as a composition of functions,
\begin{equation}
    \bm{y} = q_m(\ldots (q_2(q_1(\bm{x}))) \ldots).
\end{equation}
We can think of an event network as perturbing the output of each $q_i$ by some $\bm{\epsilon}_i$. That is, 
\begin{equation}
    \bm{y}' = q_m(\ldots (q_2(q_1(x) + \bm{\epsilon}_1) + \bm{\epsilon}_2) \ldots) + \bm{\epsilon}_m.
\end{equation}
If we assume a threshold policy with threshold $h$, then $\|\bm{\epsilon}_i\|_\infty < h$. Given these facts and some knowledge of the properties of the $q_i$ (e.g., the distribution of their weights), can we bound the norm of $\bm{y} - \bm{y}'$? This question has important implications for applications that require accuracy guarantees and should be studied in future work.


\clearpage

\begin{table}
    \centering
    \caption{\textbf{Video Pose Estimation.} Detailed results for the OpenPose model on JHMDB. The ``Skip-Conv Reset'' model re-flushes the network (i.e., sets all thresholds to zero) every \num{8} frames. See Fig.~7.}
    \label{tab:video_pose_estimation}
    \begin{tabular}{lccc}
        \toprule
        Model           & Threshold  & PCK          & Operations     \\ \midrule
        Conventional    & --         & \num{0.7640} & \num{7.055e10} \\
        EvNet           & \num{0.01} & \num{0.7656} & \num{1.071e10} \\
        EvNet           & \num{0.02} & \num{0.7696} & \num{8.825e9}  \\
        EvNet           & \num{0.04} & \num{0.7637} & \num{6.780e9}  \\
        EvNet           & \num{0.06} & \num{0.7448} & \num{5.640e9}  \\
        EvNet           & \num{0.08} & \num{0.7310} & \num{4.890e9}  \\
        Skip-Conv       & \num{0.01} & \num{0.7603} & \num{1.027e10} \\
        Skip-Conv       & \num{0.02} & \num{0.7277} & \num{7.873e9}  \\
        Skip-Conv       & \num{0.04} & \num{0.6644} & \num{5.837e9}  \\
        Skip-Conv Reset & \num{0.01} & \num{0.7621} & \num{1.092e10} \\
        Skip-Conv Reset & \num{0.02} & \num{0.7311} & \num{8.816e9}  \\
        Skip-Conv Reset & \num{0.04} & \num{0.6635} & \num{7.054e9}  \\
        \bottomrule
    \end{tabular}
\end{table}

\begin{table}
    \centering
    \caption{\textbf{Video Object Detection.} Detailed results for the YOLO model on VID. The ``Skip-Conv Reset'' model re-flushes the network (i.e., sets all thresholds to zero) every \num{8} frames. See Fig.~7.}
    \label{tab:video_object_detection}
    \begin{tabular}{lccc}
        \toprule
        Model           & Threshold  & mAP50        & Operations     \\ \midrule
        Conventional    & --         & \num{0.5538} & \num{1.537e10} \\
        EvNet           & \num{0.01} & \num{0.5545} & \num{7.472e9}  \\
        EvNet           & \num{0.02} & \num{0.5563} & \num{6.074e9}  \\
        EvNet           & \num{0.04} & \num{0.5618} & \num{4.517e9}  \\
        EvNet           & \num{0.08} & \num{0.5619} & \num{3.061e9}  \\
        EvNet           & \num{0.12} & \num{0.5463} & \num{2.306e9}  \\
        EvNet           & \num{0.16} & \num{0.5024} & \num{1.812e9}  \\
        Skip-Conv       & \num{0.01} & \num{0.5413} & \num{7.340e9}  \\
        Skip-Conv       & \num{0.02} & \num{0.4581} & \num{5.705e9}  \\
        Skip-Conv       & \num{0.04} & \num{0.3098} & \num{3.819e9}  \\
        Skip-Conv Reset & \num{0.01} & \num{0.5406} & \num{8.111e9}  \\
        Skip-Conv Reset & \num{0.02} & \num{0.4544} & \num{6.692e9}  \\
        Skip-Conv Reset & \num{0.04} & \num{0.2737} & \num{5.054e9}  \\
        \bottomrule
    \end{tabular}
\end{table}

\begin{table}
    \centering
    \caption{\textbf{Operation Overhead.} The amount of overhead operations required for computing EvNet updates. ``Math'' refers to arithmetic operations (additions and subtractions) and ``load/store'' refers to memory access operations. Percentages are the ratio of additional operations expended for each arithmetic operation saved. For example, a memory overhead of \SI{1}{\percent} indicates that one extra load/store is expended for each \num{100} arithmetic operations saved. See Table~2.}
    \label{tab:operation_overhead}
    \begin{tabular}{lccc}
        \toprule
        Model    & Threshold  & Load/Store          & Memory              \\ \midrule
        OpenPose & \num{0.01} & \SI{0.16}{\percent} & \SI{0.28}{\percent} \\
        OpenPose & \num{0.02} & \SI{0.14}{\percent} & \SI{0.25}{\percent} \\
        OpenPose & \num{0.04} & \SI{0.12}{\percent} & \SI{0.21}{\percent} \\
        OpenPose & \num{0.06} & \SI{0.10}{\percent} & \SI{0.18}{\percent} \\
        OpenPose & \num{0.08} & \SI{0.09}{\percent} & \SI{0.16}{\percent} \\
        YOLO     & \num{0.01} & \SI{0.88}{\percent} & \SI{1.57}{\percent} \\
        YOLO     & \num{0.02} & \SI{0.74}{\percent} & \SI{1.28}{\percent} \\
        YOLO     & \num{0.04} & \SI{0.62}{\percent} & \SI{1.04}{\percent} \\
        YOLO     & \num{0.08} & \SI{0.52}{\percent} & \SI{0.85}{\percent} \\
        YOLO     & \num{0.12} & \SI{0.47}{\percent} & \SI{0.74}{\percent} \\
        YOLO     & \num{0.16} & \SI{0.43}{\percent} & \SI{0.67}{\percent} \\
        \bottomrule
    \end{tabular}
\end{table}

\begin{table}
    \centering
    \caption{\textbf{Video Pose Estimation for Larger Images.}  The corrolary of \autoref{tab:video_pose_estimation}, but for larger input images ($\num{352} \times \num{480}$ instead of $\num{320} \times \num{240}$).}
    \label{tab:video_pose_estimation_for_larger_images}
    \begin{tabular}{lccc}
        \toprule
        Model           & Threshold  & PCK          & Operations     \\ \midrule
        Conventional    & --         & \num{0.8181} & \num{1.591e11} \\
        EvNet           & \num{0.01} & \num{0.8171} & \num{2.324e10} \\
        EvNet           & \num{0.02} & \num{0.8200} & \num{1.903e10} \\
        EvNet           & \num{0.04} & \num{0.8073} & \num{1.446e10} \\
        EvNet           & \num{0.06} & \num{0.7785} & \num{1.196e10} \\
        EvNet           & \num{0.08} & \num{0.7489} & \num{1.033e10} \\
        Skip-Conv       & \num{0.01} & \num{0.8065} & \num{2.183e10} \\
        Skip-Conv       & \num{0.02} & \num{0.7656} & \num{1.660e10} \\
        Skip-Conv       & \num{0.04} & \num{0.6852} & \num{1.212e10} \\
        Skip-Conv Reset & \num{0.01} & \num{0.8084} & \num{2.344e10} \\
        Skip-Conv Reset & \num{0.02} & \num{0.7689} & \num{1.886e10} \\
        Skip-Conv Reset & \num{0.04} & \num{0.6940} & \num{1.497e10} \\
        \bottomrule
    \end{tabular}
\end{table}

\begin{table}
    \centering
    \caption{\textbf{Video Object Detection for Larger Images.} The corrolary of \autoref{tab:video_object_detection}, but for larger input images ($\num{320} \times \num{544}$ instead of $\num{224} \times \num{384}$).}
    \label{tab:video_object_detection_for_larger_images}
    \begin{tabular}{lccc}
        \toprule
        Model           & Threshold  & mAP50        & Operations     \\ \midrule
        Conventional    & --         & \num{0.5655} & \num{3.164e10} \\
        EvNet           & \num{0.01} & \num{0.5658} & \num{1.527e10} \\
        EvNet           & \num{0.02} & \num{0.5679} & \num{1.246e10} \\
        EvNet           & \num{0.04} & \num{0.5726} & \num{9.289e9}  \\
        EvNet           & \num{0.08} & \num{0.5785} & \num{6.284e9}  \\
        EvNet           & \num{0.12} & \num{0.5616} & \num{4.714e9}  \\
        EvNet           & \num{0.16} & \num{0.5007} & \num{3.686e9}  \\
        Skip-Conv       & \num{0.01} & \num{0.5525} & \num{1.498e10} \\
        Skip-Conv       & \num{0.02} & \num{0.4716} & \num{1.164e10} \\
        Skip-Conv       & \num{0.04} & \num{0.3140} & \num{7.726e09} \\
        Skip-Conv Reset & \num{0.01} & \num{0.5552} & \num{1.656e10} \\
        Skip-Conv Reset & \num{0.02} & \num{0.4624} & \num{1.366e10} \\
        Skip-Conv Reset & \num{0.04} & \num{0.2829} & \num{1.026e10} \\
        \bottomrule
    \end{tabular}
\end{table}

\begin{table}
    \centering
    \caption{\textbf{Operation Overhead for Larger Images.} The corrolary of \autoref{tab:operation_overhead}, but for larger input images ($\num{352} \times \num{480}$ for OpenPose and $\num{320} \times \num{544}$ for YOLO).}
    \label{tab:operation_overhead_for_larger_images}
    \begin{tabular}{lccc}
        \toprule
        Model    & Threshold  & Load/Store          & Memory              \\ \midrule
        OpenPose & \num{0.01} & \SI{0.15}{\percent} & \SI{0.26}{\percent} \\
        OpenPose & \num{0.02} & \SI{0.13}{\percent} & \SI{0.23}{\percent} \\
        OpenPose & \num{0.04} & \SI{0.11}{\percent} & \SI{0.19}{\percent} \\
        OpenPose & \num{0.06} & \SI{0.09}{\percent} & \SI{0.16}{\percent} \\
        OpenPose & \num{0.08} & \SI{0.08}{\percent} & \SI{0.14}{\percent} \\
        YOLO     & \num{0.01} & \SI{0.85}{\percent} & \SI{1.51}{\percent} \\
        YOLO     & \num{0.02} & \SI{0.71}{\percent} & \SI{1.23}{\percent} \\
        YOLO     & \num{0.04} & \SI{0.60}{\percent} & \SI{1.00}{\percent} \\
        YOLO     & \num{0.08} & \SI{0.50}{\percent} & \SI{0.81}{\percent} \\
        YOLO     & \num{0.12} & \SI{0.44}{\percent} & \SI{0.71}{\percent} \\
        YOLO     & \num{0.16} & \SI{0.40}{\percent} & \SI{0.64}{\percent} \\
        \bottomrule
    \end{tabular}
\end{table}

\clearpage
\bibliographystyle{splncs04}
\bibliography{references}